\documentclass{article}
\usepackage{iftex}
\ifPDFTeX\pdfminorversion=7\fi
\ifXeTeX\fi
\PassOptionsToPackage{square,comma,numbers,sort&compress}{natbib}
\usepackage[preprint]{neurips_2026}
\makeatletter\def\@trackname{}\makeatother

\usepackage[utf8]{inputenc}
\usepackage[T1]{fontenc}
\usepackage[table,dvipsnames]{xcolor}
\definecolor{cvprblue}{rgb}{0.21,0.49,0.74}
\usepackage[pdfversion=1.7,breaklinks,colorlinks,citecolor=cvprblue]{hyperref}
\hypersetup{urlcolor=cvprblue,linkcolor=cvprblue}
\usepackage{xurl}
\usepackage{booktabs}       %
\usepackage{amsfonts}       %
\usepackage{nicefrac}
\usepackage{multirow}
\usepackage{microtype}
\usepackage{xcolor}
\usepackage{graphicx}
\usepackage{tabularx}
\usepackage{amsmath}        %
\usepackage{amssymb}        %
\usepackage{dsfont}         %
\usepackage{amsthm}         %
\usepackage{subcaption}
\usepackage[export]{adjustbox}  %
\usepackage{enumitem}
\theoremstyle{plain}
\newtheorem{theorem}{Theorem}
\newtheorem{lemma}[theorem]{Lemma}
\newtheorem{corollary}[theorem]{Corollary}
\newtheorem{proposition}[theorem]{Proposition}
\theoremstyle{definition}
\newtheorem{definition}{Definition}
\newtheorem{assumption}{Assumption}
\theoremstyle{remark}
\newtheorem{remark}{Remark}
\newcommand{\ind}{\mathds{1}}

\title{Understanding the Out-of-Distribution Generalization of Chain-of-Thought Reasoning in LLMs}

\author{
  Nuo-Jing Liang\textsuperscript{1}\quad
  Xiao-Tong Yuan\textsuperscript{1}\thanks{Corresponding author.}\quad
  Pan Zhou\textsuperscript{2}\\[0.15em]
  {\small\textsuperscript{1}School of Intelligence Science and Technology, Nanjing University\quad
   \textsuperscript{2}Singapore Management University}\\[0.05em]
  {\small\texttt{nuojingliang@smail.nju.edu.cn,\ \ xtyuan@nju.edu.cn,\ \ panzhou@smu.edu.sg}}
}

\hypersetup{
  linkcolor=black,
  urlcolor=cvprblue,
  pdftitle={Understanding the Out-of-Distribution Generalization of Chain-of-Thought Reasoning in LLMs},
  pdfauthor={Nuo-Jing Liang, Xiao-Tong Yuan, Pan Zhou}
}
\allowdisplaybreaks[1]

\begin{document}

\maketitle

\begin{abstract}
Chain-of-Thought (CoT) reasoning substantially improves Large Language Models (LLMs), yet its
out-of-distribution (OOD) generalization mechanism remains theoretically underexplored. We study
CoT as compositional OOD generalization: the target task lacks complete reasoning trajectories or
direct input-output pairs, while training provides only in-distribution (ID) atomic step data. We
show that CoT generalizes by repeatedly applying a shared single-step predictor to compose reusable
ID transitions. Theoretically, let $T$ denote the number of reasoning steps in the target task, $n$
the number of training samples per sub-task, and $k$ the number of output tokens per step; then the
OOD task risk is controlled by the sum of the ID sub-task risks, scaled by a coefficient of order $\mathcal{O}(k)$ that is independent of $T$, although the summed risk itself may still grow with $T$. A
Rademacher-complexity analysis further shows that the estimation term is of order $\mathcal{O}(\sqrt{T/n})$,
rather than linear in $T$, due to predictor sharing under a uniform capacity control on the shared
class. For autoregressive Next-Token Prediction, the final-task risk is bounded by the best-in-class step-level risk scaled by the same coefficient of order $\mathcal{O}(k)$, plus an estimation term of order $\widetilde{\mathcal{O}}\big(k\sqrt{T/n}\big)$. A parity-coupled construction further shows that, for the sub-task empirical risk minimizer, the expected final-task risk is at least the best in-class risk plus $\Omega(\sqrt{T/n})$ when $T\le n$. Experiments on Atom-Task and GSM8K support the theory: CoT enables
short-to-long and compositional generalization, while explicit state tracking is crucial.

Our code is available at \href{https://github.com/maxliang114514/CoT-OOD}{\nolinkurl{https://github.com/maxliang114514/CoT-OOD}}.
\end{abstract}

\section{Introduction}
\label{sec:introduction}
Chain-of-Thought (CoT) reasoning has become a central mechanism for improving the reasoning ability of Large Language Models (LLMs). By prompting or training models to generate intermediate steps, CoT substantially improves mathematical and logical reasoning~\cite{wei2022chain,yuemammoth,yumetamath,wang2024math,shao2024deepseekmath,yu2024flow,ho2023large,kim2023cot}. Recent reasoning-oriented systems further treat step generation as an explicit training signal~\cite{openai2024learning,guo2025deepseek,gandhicognitive}, shifting LLM reasoning from direct pattern matching toward structured stepwise computation~\cite{prabhakar2024deciphering}. Despite this empirical success, a fundamental theoretical question remains open: \emph{why can CoT-trained models generalize to complex reasoning tasks that are not directly observed during training?}

Existing theoretical studies provide important insights into the expressivity and computational role of CoT. Some works show that CoT can increase the effective depth of Transformers or enable serial computation~\cite{feng2023towards,merrillexpressive,li2024chain}, while others analyze task decomposition, in-context learning, or localized reasoning settings~\cite{yang2025chain,bavandpour2025lower,prystawski2023think}. However, most existing analyses are either qualitative, restricted to narrow or two-step settings, or do not provide explicit generalization bounds for the autoregressive Next-Token Prediction (NTP) regime commonly used by modern large-scale LLMs. As a result, the out-of-distribution (OOD) generalization mechanism of multi-step CoT remains theoretically underexplored.

\begin{figure}[htbp]
	\centering
	\includegraphics[width=\linewidth]{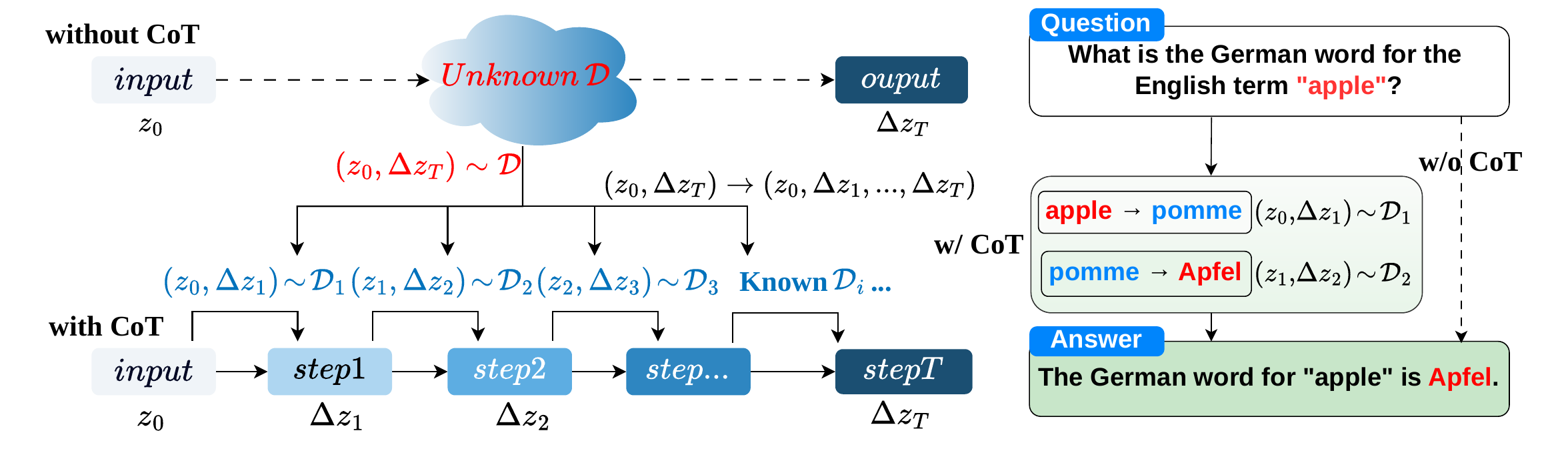}
	\caption{
		CoT as compositional OOD generalization.
		Left: without CoT, the model must learn a direct mapping from input $z_0$ to final answer $\Delta z_T$, whose target distribution may be unavailable during training.
		With CoT, the same predictor is repeatedly applied to in-distribution step distributions, mapping each state $z_{i-1}$ to update $\Delta z_i$.
		Right: a two-hop translation example solves an unseen English-to-German query by composing learned English-to-French and French-to-German steps.
	}
	\label{fig:fig1}
\end{figure}

 This paper studies the mechanism of CoT through compositional OOD generalization. We consider a target task requiring a complex multi-step mapping from an input state $z_0$ to a final answer $\Delta z_T$, while neither complete reasoning trajectories $(z_0,\Delta z_1,\ldots,\Delta z_{T-1},\Delta z_T)$ nor direct input-output pairs $(z_0,\Delta z_T)$ are available during training. Thus, the target task's composition is OOD relative to the observed atomic transitions. We refer to this setting as compositional OOD; Section~\ref{sec:rademacher} separately treats mismatched training and target step distributions (Corollary~\ref{cor:atomic_shift}). Instead, the model has access to in-distribution (ID)  atomic step data: samples $(z_{i-1},\Delta z_i)$, where $z_{i-1}$ is the reasoning state before step $i$ and $\Delta z_i$ is the next incremental output. Here, $T$ denotes the total number of generation steps: the first $T-1$ are intermediate reasoning updates, and $\Delta z_T$ is the final answer.

Under this view, CoT does not require the model to learn the entire OOD mapping at once from scratch. Instead, it repeatedly applies a shared single-step predictor. At each step, the model reads the current state $z_{i-1}$, generates $\Delta z_i$ containing $k$ new tokens, and updates the state by appending these tokens while keeping a fixed context length $m$. CoT therefore converts a complex target mapping into a composition of reusable ID atomic transitions (Fig.~\ref{fig:fig1}). This motivates our central question:

\begin{center}
\emph{When a model learns only atomic step transitions, can repeated CoT composition generalize to unseen 
multi-step OOD tasks, and how does the resulting error scale with the reasoning depth $T$?}
\end{center}

\paragraph{Contributions} We answer this question by developing a generalization framework for CoT-based task decomposition. First, we prove that the risk of the complex OOD task can be upper-bounded by the sum of risks over ID sub-tasks, scaled by a coefficient of order $\mathcal{O}(k)$ that is independent of $T$. The coefficient results from deterministic argmax decoding together with the bounded sequence loss and the local error floor $(\log 2)/k$ that a hard-decoding (argmax) error induces. CoT thus avoids fitting the unseen composite mapping directly and composes local transitions at test time.

Second, we derive a Rademacher-complexity bound for the multi-step setting. A naive analysis that treats each reasoning step independently would yield a linear dependence on $T$. In contrast, because CoT applies the same predictor across all steps, we can analyze the combined step-level sample set and obtain a sublinear generalization-gap term scaling as $\mathcal{O}(\sqrt{T/n})$, where $n$ is the number of training samples per atomic step distribution.

Third, we specialize the analysis to the NTP setting of autoregressive LLMs. We directly analyze the final $k$ target tokens of each atomic sequence and take the $N=nT$ complete sequences as the sampling units. The resulting final-task risk bound keeps the best in-class step-level risk explicit and adds an estimation term of order $\widetilde{\mathcal{O}}\big(k\sqrt{T/n}\big)$. Appendix~\ref{app:proof_pooled_lower} shows that the $\sqrt{T/n}$ dependence cannot be improved for the sub-task empirical risk minimizer: a construction with $k=1$, in which the per-step errors are coupled through the XOR accumulated over the chain, makes this minimizer incur an excess of $\Omega(\sqrt{T/n})$ when $T\le n$, and carries that excess through to the final task.

We further validate our theory through controlled experiments. We introduce Atom-Task, a symbolic compositional benchmark involving multi-step sequence operations, to isolate the effect of CoT on OOD compositional generalization. Experiments show that CoT-trained models generalize from shorter or constrained training chains to longer unseen compositions, whereas models trained without intermediate reasoning fail to capture the underlying compositional structure. Ablation results further show that explicitly tracking the current reasoning state is crucial, matching our theoretical requirement that each state $z_{i-1}$ contains sufficient information for the next atomic transition. Additional experiments on GSM8K, a benchmark of grade-school math word problems, support the same conclusion in real-world mathematical reasoning.

\section{Preliminaries}
\label{pre}

\paragraph{Problem Setting}
We consider an autoregressive predictor with a context window size of $m$. Let $\mathcal{V}$ be a vocabulary (a set of tokens), with $n_v = |\mathcal{V}|$ denoting its size, and let $\mathcal{V}^m$ denote the set of all token sequences of length $m$ formed by tokens from $\mathcal{V}$. The predictor $\psi: \mathcal{V}^m \to \mathbb{R}^{n_v \times k}$ ($m \gg k$) maps an input sequence to probability distributions over $k$ output positions, and belongs to a hypothesis class $\Psi$. Losses are computed on these soft predictions; for autoregressive generation, discrete tokens are obtained via argmax, denoted $\mathrm{hard}(\psi(\cdot)) \in \mathcal{V}^k$.

Under the CoT paradigm, the model does not predict the final answer directly via $\psi(z_0)$, but generates a chain of $T$ reasoning steps. We write $\widehat{\cdot}$ for model-generated quantities and reserve the unhatted symbol for the ground truth. At step $i$, the incremental output is $\widehat{\Delta z}_i$ $= \mathrm{hard}(\psi(\widehat z_{i-1}))$, and the state updates by appending these $k$ tokens while discarding the first $k$ (sliding window):
\begin{equation*}
    \widehat z_i = \widehat z_{i-1}[k{+}1:m] \oplus \widehat{\Delta z}_i \quad (\oplus \text{ denotes sequence concatenation}), \quad \widehat{\Delta z}_T = \mathrm{hard}(\psi(\widehat z_{T-1})),
\end{equation*}
thus maintaining the sequence length $m$ for all intermediate states, with $\widehat z_0=z_0$; the true state follows the same recursion with $\Delta z_i$ in place of $\widehat{\Delta z_i}$, i.e., $z_i=z_{i-1}[k+1:m]\oplus\Delta z_i$.

For a given complex task, any instance $(z_0, \Delta z_T)$ with $z_0 \in \mathcal{V}^m$ and $\Delta z_T \in \mathcal{V}^k$ satisfies $(z_0, \Delta z_T) \sim \mathcal{D}$. There exists a ground-truth mapping $\psi_{*}$ (not necessarily within $\Psi$) such that $\mathrm{hard}(\psi_{*}(z_{i-1})) = \Delta z_i$ for all $i$, implying $\psi_{*}$ generates the CoT sequence $(\Delta z_1, \dots, \Delta z_T)$ from $z_0$ to yield the target output $\Delta z_T$.

Define the state transition $\bar{\psi}: \mathcal{V}^m \to \mathcal{V}^m$ as $\bar{\psi}(z) := z[k{+}1:m] \oplus \mathrm{hard}(\psi(z))$, representing the input state for the next step (note $\bar\psi$ uses hard predictions for transitions). To distinguish from $\psi$, let $F_\psi: \mathcal{V}^m \to \mathbb{R}^{n_v \times k}$ denote the composite CoT mapping $F_\psi(z_0) := \psi(\bar{\psi}^{(T-1)}(z_0))$, which encapsulates the $T$-step reasoning process (note $F_\psi$ applies $\psi$ to the final state, outputting soft predictions for loss computation). Our learning goal is to minimize the population risk:
\begin{equation*}
    R_{\mathcal{D}}(F_\psi) = \int \mathfrak{L} (F_\psi(z_0), \Delta z_T) \, dP(z_0, \Delta z_T) = \int \mathfrak{L} (\psi(\bar{\psi}^{(T-1)}(z_0)), \Delta z_T) \, dP(z_0, \Delta z_T),
\end{equation*}
where $P$ is the probability measure associated with $\mathcal D$ and $\mathfrak{L}: \mathbb{R}^{n_v \times k} \times \mathbb{R}^{n_v \times k} \to \mathbb{R}$ is the sequence-level loss function that compares the model's soft predictions (in $\mathbb{R}^{n_v \times k}$) with the ground-truth output $\Delta z_T$ (converted from discrete tokens to one-hot vectors in $\mathbb{R}^{n_v \times k}$); the formal definition is provided in Section~\ref{sec:theoretical_analysis}. This aims to identify a $\psi \in \Psi$ approximating $\psi_*$, thereby generating the CoT sequence from $z_0$ to yield $\Delta z_T$.

\paragraph{Notations}
$\bar\psi^{(T)}$ denotes the $T$-fold composition $\bar\psi \circ \cdots \circ \bar\psi$ ($T$ times in total). The final prediction is extracted via $\psi(\bar\psi^{(T-1)}(z_0))$. For a sequence $z$, $z[a:b]$ denotes the subsequence from index $a$ to $b$ (inclusive), and $(z)_j$ denotes its $j$-th component (or token). The operator $\oplus$ denotes the concatenation of two sequences. Other key notations are summarized in Table~\ref{tab:notation} in Appendix~B.

\section{Theoretical Analysis: Task Decomposition for OOD Tasks}
\label{sec:theoretical_analysis}

We analyze the generalization error of CoT reasoning by decomposing the risk of the complex OOD task into per-step risks (Section~\ref{sec:decompose}) and then bounding the generalization gap  (Section~\ref{sec:rademacher}).

\begin{assumption}[Bounded and Lipschitz Loss Function]
\label{assum:lipschitz}
For a single token represented as a probability vector in $\mathbb{R}^{n_v}$, the token-level loss function $\ell(\cdot,\cdot)$ is defined as the $\epsilon_\ell$-clipped cross-entropy loss with a fixed clipping threshold $\epsilon_\ell\in(0,1/4)$, i.e., $\ell(p,y):=-\log\max\{p_y,\ \epsilon_\ell\}$, where $p$ denotes the predicted probability vector and $y$ the ground-truth token. It is upper-bounded by $B_\ell$, i.e., $\ell(p,y)\le B_\ell$, and is $G_\ell$-Lipschitz continuous with respect to the $\ell_2$-norm of the predicted probability vector, i.e., $\left|\ell(\hat{y}_1, y) - \ell(\hat{y}_2, y)\right| \le G_\ell \|\hat{y}_1 - \hat{y}_2\|_2$ for all probability vectors $\hat{y}_1,\hat{y}_2\in\Delta_{n_v-1}:=\{p\in[0,1]^{n_v}:\sum_{a=1}^{n_v}p_a=1\}$ and ground-truth token $y$. Without loss of generality, assume $G_\ell \ge 1$.
\end{assumption}

\begin{remark}
\label{rem:assumptions}
Assumption~\ref{assum:lipschitz} is standard in learning theory~\cite{shalev2014understanding, liu2022gradient,deng2024generalization,liu2025consistency}. Both requirements follow directly from clipping, which lower-bounds the probability assigned to the true token by $\epsilon_\ell$; both conditions are therefore satisfied by construction.
\end{remark}

\begin{definition}[Sequence Loss]
For a predicted probability distribution $\hat{y} \in \mathbb{R}^{n_v \times k}$ and a ground-truth token sequence $\Delta z$ (represented as one-hot vectors in $\mathbb{R}^{n_v \times k}$), the sequence loss is defined as the average of the token-wise loss $\ell$ over these $k$ generated tokens, where $\hat{y}_{j}$ denotes the probability vector at output position $j$ (the $j$-th column of $\hat y$) \cite{li2025generalization}:
\begin{equation*}
    \mathfrak{L} (\hat{y}, \Delta z) := \frac{1}{k} \sum\nolimits_{j=1}^{k} \ell(\hat{y}_{j}, (\Delta z)_{j})
\end{equation*}
\end{definition}

\subsection{Generalization Error Decomposition via CoT}
\label{sec:decompose}
For $(z_0,\Delta z_T)\sim \mathcal{D}$, let $\mathcal{D}'$ denote the joint distribution over the reasoning sequence $(z_0, \Delta z_1, \dots, \Delta z_T)$, and let $\mathcal D_i$ denote the marginal law of $(z_{i-1},\Delta z_i)$ under $\mathcal D'$ (a state on the ground-truth trajectory is called an oracle state). Lemma~\ref{lem:bound_M} in Appendix~\ref{app:proof_bound_M} shows that the step-level loss of an erroneous hard-decoded block is at least $(\log 2)/k$, a floor that is independent of the reasoning depth $T$.

\begin{theorem}[Generalization Error Decomposition]
\label{thm:decomp}
Under Assumption~\ref{assum:lipschitz}, we have:
\begin{equation*}
    R_{\mathcal{D}}(F_\psi) \le \frac{B_\ell k}{\log 2}\sum\nolimits_{i=1}^{T}\int \mathfrak{L}(\psi(z_{i-1}),\Delta z_i)d\mathcal{D}_i(z_{i-1},\Delta z_i),
\end{equation*}
where $(\log 2)/k$ is the local error floor of Lemma~\ref{lem:bound_M}.
\end{theorem}

\begin{remark}[Implications of Theorem~\ref{thm:decomp}]
Theorem~\ref{thm:decomp} reveals the fundamental principle underlying CoT's OOD generalization: \emph{the generalization risk of a complex OOD task is dominated by the sum of per-step in-distribution risks}, scaled by an $\mathcal{O}(k)$ coefficient that is independent of $T$. This is qualitatively different from direct OOD learning, where one would attempt to bound the risk of the composite mapping $F_\psi$ directly---without decomposition, not only is the mapping $F_\psi$ inherently more complex, but more critically, we lack the direct samples $(z_0, \Delta z_T) \sim \mathcal{D}$, so the composite risk cannot be estimated from target data alone.
\end{remark}

Theorem~\ref{thm:decomp} bounds the target risk by the summed step-level risks, and the additive dependence on those risks is unavoidable in general. A step-level error can propagate along the chain through the state update rather than being averaged out, so the per-step risks accumulate in the final answer.

\begin{remark}[Tightness of Theorem~\ref{thm:decomp}]
\label{rem:additive_tightness}
Appendix~\ref{app:proof_additive_tightness} constructs, for $k=1$, a binary sliding-window task with a shared single-step predictor $\psi$ and a bounded sequence loss, for which
\begin{equation*}
R_{\mathcal D}(F_\psi)=\Omega\!\left(\sum_{i=1}^{T}R_{\mathcal D_i}(\psi)\right).
\end{equation*}
Consequently, the additive form of Theorem~\ref{thm:decomp} is tight up to its depth-independent coefficient: when every step-level error reaches the final answer, the linear growth in $T$ of the summed risk is attained, and no argument that charges only per-step risks can remove it.
\end{remark}

\subsection{Bounds via Rademacher Complexity}
\label{sec:rademacher}

Let $S_i=\{(z^{(j)}_{i-1},\Delta z^{(j)}_i)\}_{j=1}^{n}$ be $n$ independent atomic pairs drawn from $\mathcal D_i$, with $S_0=\bigcup_{i=1}^{T}S_i$. All $nT$ atomic examples are mutually independent, with the $n$ examples in $S_i$ drawn from $\mathcal D_i$; each comes from some source realizing the same step-level skill distribution and need not lie on a common trajectory. Define $f_\psi(z,\Delta z):=\mathfrak{L}(\psi(z),\Delta z)$ and $\mathcal F:=\{f_\psi:\psi\in\Psi\}$. To bound the generalization gap we use Rademacher complexity, writing $R_{\mathcal D_i}(\psi)$ and $\widehat R_{S_i}(\psi)$ for the associated population and empirical risks (Appendix~\ref{app:proof_rad}). Applying Rademacher bounds to the $T$ steps separately makes the resulting gap grow linearly in $T$. Since the same learner $\psi$ processes samples from all $T$ distributions, analyzing the combined set $S_0$ under a uniform capacity control makes both the deviation term and the complexity term (from $\sum_{i=1}^{T}\widehat{\mathfrak R}_{S_i}(\mathcal F)$ to $T\widehat{\mathfrak R}_{S_0}(\mathcal F)$) improve from $\mathcal{O}(T)$ to $\mathcal{O}(\sqrt T)$---analogous to multi-task learning, where sharing a hypothesis across related tasks improves generalization~\cite{maurer2016benefit}. Extending the classical symmetrization lemma~\cite{shalev2014understanding} to multiple distributions (Lemma~\ref{lem:symm} in Appendix~\ref{app:proof_rad}) gives:

\begin{theorem}[Rademacher Complexity Bound]
\label{lem:rad}
For any $\delta\in(0,1)$, with probability at least $1-\delta$, uniformly over $\psi\in\Psi$, we have:
\begin{equation*}
\left|\sum\nolimits_{i=1}^{T} (R_{\mathcal{D}_i}(\psi)-\widehat{R}_{S_i}(\psi))\right| \leq 2T \widehat{\mathfrak{R}}_{S_0}(\mathcal{F}) + 3B_\ell\sqrt{\frac{2T\log(4/\delta)}{n}},
\end{equation*}
where $\widehat{\mathfrak{R}}_{S_0}(\cdot)$ denotes the empirical Rademacher complexity on $S_0$, $\mathcal F=\mathfrak{L}\circ\Psi$ is the induced bounded loss class, and $B_\ell$ is the bound of the sequence loss $\mathfrak{L}$ according to Assumption~\ref{assum:lipschitz}.
\end{theorem}

\begin{remark}
\label{rem:sqrtT}
Sharing a single learner $\psi$ across all $T$ reasoning steps, under a uniform capacity control on the shared class, is what permits treating the $T$ step-level datasets as one combined sample $S_0$ of size $nT$: the same predictor is fitted to all steps, so the estimation dependence improves from $T$ to $\sqrt{T}$.
\end{remark}

The following corollary records the resulting guarantee, first for a fixed predictor and then, after the ERM comparison, for the empirical risk minimizer; both proofs are in Appendix~\ref{app:proof_ood_bound}.

\begin{corollary}[Generalization Bound for Complex OOD Tasks]
\label{cor:ood_bound}
Under Assumption~\ref{assum:lipschitz}, with probability at least $1-\delta$, for every $\psi\in\Psi$,
\begin{equation*}
R_{\mathcal{D}}(F_\psi) \le \frac{B_\ell k}{\log 2} \Bigl[ \sum\nolimits_{i=1}^T \widehat{R}_{S_i}(\psi) + 2T \widehat{\mathfrak{R}}_{S_0}(\mathcal{F}) + 3B_\ell\sqrt{\frac{2T\log(4/\delta)}{n}} \Bigr],
\end{equation*}
and for the sub-task ERM $\widehat\psi\in\arg\min_{\psi\in\Psi}\sum\nolimits_{i=1}^{T}\widehat R_{S_i}(\psi)$, with $\psi^\star\in\arg\min_{\psi\in\Psi}\sum\nolimits_{i=1}^{T}R_{\mathcal D_i}(\psi)$,
\begin{equation*}
R_{\mathcal{D}}(F_{\widehat\psi}) \le \frac{B_\ell k}{\log 2} \Bigl[ \sum\nolimits_{i=1}^T R_{\mathcal D_i}(\psi^\star) + 4T \widehat{\mathfrak{R}}_{S_0}(\mathcal{F}) + 6B_\ell\sqrt{\frac{2T\log(4/\delta)}{n}} \Bigr].
\end{equation*}
\end{corollary}

\begin{remark}[Tightness of Corollary~\ref{cor:ood_bound}]
\label{rem:atomic_sharpness}
The estimation term of Theorem~\ref{lem:rad} is order-tight, and its rate is inherited by Corollary~\ref{cor:ood_bound}: Appendix~\ref{app:proof_pooled_lower} constructs a binary compositional task and a two-element class for which the sub-task ERM $\widehat\psi$ of Corollary~\ref{cor:ood_bound} satisfies
\begin{equation*}
\mathbb E_S\!\left[\sum_{i=1}^{T}R_{\mathcal D_i}(\widehat\psi)-\min_{\psi\in\Psi}\Bigl\{\sum_{i=1}^{T}R_{\mathcal D_i}(\psi)\Bigr\}\right]
=\Omega\!\left(\min\left\{1,\sqrt{\frac{T}{n}}\right\}\right),
\end{equation*}
i.e., its \emph{atomic excess}, the gap between its summed step-level risk and the best in-class value of that sum, is of that order; and in that instance the atomic excess coincides with the excess on the final task, so the same bound holds for $\mathbb E_S[R_{\mathcal D}(F_{\widehat\psi})-\min_{\psi\in\Psi}R_{\mathcal D}(F_\psi)]$. In particular it is $\Omega(\sqrt{T/n})$ when $T\le n$. Two features of the construction are responsible: the per-step separation is amplified by a factor of $T$ at the final token, and the best-in-class per-step risk is $\Theta(1)$, which makes the excess the identifiable quantity.
\end{remark}

The preceding analysis addresses compositional OOD generalization: the target task is out of distribution, whereas the step-level distributions $\mathcal D_i$ are not. When the step-level data are themselves drawn from shifted distributions $Q_i$, the decomposition carries over with one additional per-step term.

\begin{corollary}[Atomic Distribution Shift]
\label{cor:atomic_shift}
Let $Q_i$ be the training distribution at step $i$, let $\mathcal D_i$ be the corresponding target oracle-state marginal, and let $S_i\sim Q_i^n$. Define
\begin{equation*}
\operatorname{disc}_{\mathcal F}(\mathcal D_i,Q_i)
:=\sup_{\psi\in\Psi}
\left|R_{\mathcal D_i}(\psi)-R_{Q_i}(\psi)\right|.
\end{equation*}
Then, with probability at least $1-\delta$, for every $\psi\in\Psi$,
\begin{equation*}
R_{\mathcal D}(F_\psi)\le \frac{B_\ell k}{\log 2}\left[\sum_{i=1}^{T}\widehat R_{S_i}(\psi)+2T\widehat{\mathfrak R}_{S_0}(\mathcal F)+3B_\ell\sqrt{\frac{2T\log(4/\delta)}{n}}+\sum_{i=1}^{T}\operatorname{disc}_{\mathcal F}(\mathcal D_i,Q_i)\right],
\end{equation*}
and for the sub-task ERM $\widehat\psi\in\arg\min_{\psi\in\Psi}\sum_{i=1}^{T}\widehat R_{S_i}(\psi)$, with $\psi^\star\in\arg\min_{\psi\in\Psi}\sum_{i=1}^{T}R_{\mathcal D_i}(\psi)$,
\begin{equation*}
R_{\mathcal D}(F_{\widehat\psi})\le \frac{B_\ell k}{\log 2}\left[\sum_{i=1}^{T}R_{\mathcal D_i}(\psi^\star)+4T\widehat{\mathfrak R}_{S_0}(\mathcal F)+6B_\ell\sqrt{\frac{2T\log(4/\delta)}{n}}+2\sum_{i=1}^{T}\operatorname{disc}_{\mathcal F}(\mathcal D_i,Q_i)\right].
\end{equation*}
\end{corollary}
The proof is given in Appendix~\ref{app:proof_shift}. The corollary thus controls shifts between training and target step distributions; the drift of the model's own generated trajectory (its rollout) is handled only through the hard-decoding reduction and is not itself a distributional guarantee.

\section{Specific Analysis: Results in the Next-Token Prediction Setting}
\label{sec:ntp}

In the preceding analysis the OOD generalization error of complex compositional tasks is upper-bounded by a sum of per-step in-distribution risks, with the gap governed by the Rademacher complexity of the shared hypothesis class. We now specialize this decomposition to the Next-Token Prediction (NTP) pretraining paradigm \cite{li2025generalization}---the standard training regime for LLMs---and derive a target-token bound with explicit depth, sample-size, and decoder-capacity dependence. Relative to the general analysis of Section~\ref{sec:rademacher} the additional ingredients are (i) a reduction from the final-task risk of free-running generation, in which the model conditions on its own outputs, to target-token risks under teacher forcing, in which every prefix is taken from the ground truth (Lemma~\ref{lem:ntp_tf_reduction}, Appendix~\ref{app:proof_ntp_tf_reduction}) and (ii) an explicit evaluation of the complexity term for the decoder (Appendix~\ref{app:dim_derivation}); the ERM comparison is the one already used in Corollary~\ref{cor:ood_bound}.

\subsection{Problem Formulation}
\label{sec:ntp_form}
The model is a composition $\mathrm{LM}=g\circ h$ (representation learner $h$ and token predictor $g$). Here, the abstract single-step mapping $\psi$ is instantiated by the block predictor $\psi_{g,h}$ induced by $k$ successive next-token predictions via $g\circ h$, and $F_{\psi_{g,h}}$ denotes the resulting $T$-step final predictor. For step $i=1,\ldots,T$, let $S_i=\{X_{ij}\}_{j=1}^n$, where each $X_{ij}=(t_{ij,1},\ldots,t_{ij,m})$ is a complete sequence. We denote the teacher-forced context from $X_{ij}$ before token $t$ by $X_{ij,<t}$ and set $N=nT$. Here and in Appendix~\ref{app:ntp_setting}, $m$ denotes the complete-record length: an $(m-k)$-token state followed by $k$ target tokens. Thus the general setup of Section~\ref{pre} is instantiated with state length $m-k$. Before each block, both oracle and generated histories retain their most recent $m-k$ tokens; within the block, the corresponding true or generated prefixes are appended autoregressively. The oracle record is therefore $X=z_{i-1}\oplus\Delta z_i$, and $\mathcal D_i$ also denotes its induced distribution. This length convention leaves the general risk bounds unchanged.

For a complete sequence $X=(t_1,\ldots,t_m)$ and its prefix $X_{<t}=(t_1,\ldots,t_{t-1})$, define the target-token loss, with $g$ applied to the hidden state at the last position of its argument:
\begin{equation}
q_{\psi_{g,h}}(X):=\frac{1}{k}\sum_{t=m-k+1}^{m}
\ell\!\left(g(h(X_{<t})),t_t\right),
\label{eq:target_token_loss}
\end{equation}
and let $\mathcal Q_k=\{q_{\psi_{g,h}}:g\in\mathcal G,\,h\in\mathcal H\}$, where $\mathcal G,\mathcal H$ are the classes of the token predictor and of the representation learner. For $S_i=\{X_{ij}\}_{j=1}^n$, define
\begin{equation*}
R_{\mathcal D_i}^{k}(\psi_{g,h})
:=\mathbb E_{X\sim\mathcal D_i}[q_{\psi_{g,h}}(X)],
\qquad
\widehat R_{S_i}^{k}(\psi_{g,h})
:=\frac{1}{n}\sum_{j=1}^{n}q_{\psi_{g,h}}(X_{ij}).
\end{equation*}
The $N=nT$ complete sequences are the independent sampling units; tokens and prefixes within each sequence may be dependent. We abbreviate $R_i^{k}:=R_{\mathcal D_i}^{k}$ and $\widehat R_i^{k}:=\widehat R_{S_i}^{k}$.

\begin{remark}[Teacher-Forced NTP and Free-Running Generation]
\label{rem:oracle_prefix}
The target-token loss in Eq.~\eqref{eq:target_token_loss} is \emph{teacher-forced}: it uses the true within-step prefixes, whereas $F_{\psi_{g,h}}$ runs freely and conditions on its own generated tokens. These block losses need not coincide. Lemma~\ref{lem:ntp_tf_reduction} in Appendix~\ref{app:proof_ntp_tf_reduction} extends the first-error argument of Theorem~\ref{thm:decomp} to individual tokens and gives $R_{\mathcal D}(F_{\psi_{g,h}})\le \frac{B_\ell k}{\log 2}\sum_{i=1}^{T}R_{\mathcal D_i}^{k}(\psi_{g,h})$ with the same coefficient $B_\ell k/\log 2=\mathcal O(k)$.
\end{remark}

In this setting we write $(\widehat g,\widehat h)$ for the minimizer of $\sum_i\widehat R_{S_i}^{k}(\psi_{g,h})$ and $(g^\star,h^\star)$ for the minimizer of $\sum_iR_{\mathcal D_i}^{k}(\psi_{g,h})$, with $\widehat\psi:=\psi_{\widehat g,\widehat h}$ and $\psi^\star:=\psi_{g^\star,h^\star}$; the resulting target-token excess is the NTP analogue of the atomic excess of Section~\ref{sec:rademacher}.

\subsection{Main Theoretical Results}

Let $\widehat{\mathfrak R}_{S_0}(\mathcal Q_k)$ denote the empirical Rademacher complexity of the target-token class on the combined sample $S_0$ of $N=nT$ records, i.e., the quantity of Appendix~\ref{app:preliminaries} with $\mathcal F$ replaced by $\mathcal Q_k$ and one independent sign per record. Applying Theorem~\ref{lem:rad} to the target-token class yields the following deviation bound, uniform over the class.

\begin{theorem}[Target-Token Risk Bound]
\label{thm:ntp}
Under Assumption~\ref{assum:lipschitz}, with probability at least $1-\delta$, uniformly over $g\in\mathcal G$ and $h\in\mathcal H$,
\begin{equation}
\left|\sum_{i=1}^{T}\left(R_{\mathcal D_i}^{k}(\psi_{g,h})-\widehat R_{S_i}^{k}(\psi_{g,h})\right)\right|
\le 2T\widehat{\mathfrak R}_{S_0}(\mathcal Q_k)+3B_\ell\sqrt{\frac{2T\log(4/\delta)}{n}}.
\label{eq:target_ntp_bound}
\end{equation}
\end{theorem}

The next corollary is the target-token counterpart of the ERM form of Corollary~\ref{cor:ood_bound}: for the empirical risk minimizer the empirical risks in the bound are replaced by the best-in-class population risks, at the price of paying the deviation twice.

\begin{corollary}[ERM Final-Task Risk]
\label{cor:erm_final_risk}
Let $(\widehat g,\widehat h)$ and $(g^\star,h^\star)$ be defined as above. With probability at least $1-\delta$,
\begin{equation}
R_{\mathcal D}(F_{\psi_{\widehat g,\widehat h}})\le \frac{B_\ell k}{\log 2}\left[\sum_{i=1}^{T}R_{\mathcal D_i}^{k}(\psi_{g^\star,h^\star})+4T\widehat{\mathfrak R}_{S_0}(\mathcal Q_k)+6B_\ell\sqrt{\frac{2T\log(4/\delta)}{n}}\right].
\label{eq:erm_final_risk}
\end{equation}
\end{corollary}
Under Assumption~\ref{assum:arch}, the target-token complexity obeys the following bound. Deriving it (Appendix~\ref{app:dim_derivation}) requires two ingredients beyond the general analysis of Section~\ref{sec:rademacher}: the layerwise matrix-covering argument of \cite{li2025generalization}, which we instantiate for our decoder block and then restrict to the final $k$ output rows, and a sequence-level loss-covering step of our own that charges one Rademacher sign per complete sequence rather than per generated token; the latter is what produces the $1/\sqrt{N}$ normalization with $N=nT$.
\begin{equation}
\widehat{\mathfrak R}_{S_0}(\mathcal Q_k)
\lesssim
G_\ell\sqrt{\frac{\Theta dH\tau_{m,N}}{N}}
+G_\ell\sqrt{\frac{dn_v}{N}},
\qquad N=nT,
\label{eq:qk_capacity}
\end{equation}
where $\Theta$ counts trainable parameters, $L,H,d$ are the layer count, head count and model dimension, and $\tau_{m,N}=\log(1+\rho_Ls_L(m,N))$ is the decoder-capacity factor (Appendix~\ref{app:dim_derivation}); below we write $\tau$ for $\tau_{m,N}$.

The resulting final-task bound is, for constants $C_1,C_2>0$ depending only on the fixed norm and normalization bounds of Assumption~\ref{assum:arch},
\begin{equation}
R_{\mathcal D}(F_{\psi_{\widehat g,\widehat h}})\le \frac{B_\ell k}{\log 2}\bigl[\sum\nolimits_{i=1}^{T}R_{i}^{k}(\psi^\star)
+\!C_1G_\ell\sqrt{\frac{T\Theta dH\tau}{n}}+C_2G_\ell\sqrt{\frac{Tdn_v}{n}}
+\!6B_\ell\sqrt{\frac{2T\log(4/\delta)}{n}}\bigr].
\label{eq:final_parameterized_ntp}
\end{equation}
Since $B_\ell k/\log 2=\mathcal{O}(k)$, for fixed model dimensions, norm bounds and clipping threshold the sample-dependent atomic excess contributes $\widetilde{\mathcal O}(k\sqrt{T/n})$ to the bound, where $\widetilde{\mathcal O}$ suppresses the logarithmic factors of $\tau$ and of the confidence term.

\subsection{Discussion of the Final Bound}
\paragraph{Depth and sample size.}
The target-token analysis treats the $N=nT$ complete sequences as the independent observations while allowing arbitrary dependence among tokens within each sequence. Pooling these observations through one shared predictor under a uniform capacity control on the class yields an atomic excess of order $\sqrt{T/n}=\sqrt{T}\cdot n^{-1/2}$---a $\sqrt{T}$ accumulation across the $T$ composed steps, sublinear in place of the naive linear $T$, times the $n^{-1/2}$ statistical cost of a single atomic step---rather than the $T/\sqrt n$ dependence obtained by analyzing the $T$ steps separately. We state this with the per-step sample size $n$ held fixed, so that the depth dependence is $\sqrt{T}$; if the total budget $N=nT$ is held fixed instead, then $n$ shrinks and the per-step sampling factor alone grows like $\sqrt{T/N}$, giving $T/\sqrt{N}$---linear behavior that comes from the shrinking per-step budget, not from any loss of the pooling gain. Equation~\eqref{eq:erm_final_risk} retains the best-in-class step-level risk in the complete final-task risk bound; after the depth-independent factor $B_\ell k/\log 2=\mathcal O(k)$, the sample-dependent atomic excess contributes $\widetilde{\mathcal O}(k\sqrt{T/n})$ to the bound.

\paragraph{Context length and decoder capacity.}
We analyze the final-$k$-token loss directly, which permits dependence among tokens within a sequence; concentration is taken over the $N=nT$ independent complete sequences. In the parameterized bound of Eq.~\eqref{eq:final_parameterized_ntp}, context length enters through $\tau$ evaluated at $N=nT$ and also through $\Theta$ if the trainable positional-embedding parameter count grows with $m$.

\paragraph{Theoretical context} Unlike prior work that restricts compositional analysis to 2-step cases without explicit parameter-dependent bounds~\cite{yao2026compositional}, 
focuses on sub-task identification sample complexity rather than OOD error decomposition~\cite{abedsoltan2025task},
proves that CoT reduces learning difficulty without generalization bounds~\cite{yang2025chain},
or addresses only single-distribution pretraining~\cite{li2025generalization},
our framework unites CoT error decomposition with a shared-hypothesis structure and directly controls target-token risk with reasoning-depth and decoder-capacity dependence. Maurer et al.~\cite{maurer2016benefit} provide a shared-representation perspective for pooled estimation, whereas Li et al.~\cite{li2025generalization} provide decoder covering-number tools for full-sequence NTP. We use the latter only to control the restricted final-$k$ output class; the target-token risk bound itself is derived directly at the sequence level.

\section{Experiments}
\label{sec:experiments}

We evaluate compositional generalization on Atom-Task, a task of multi-step operations on sequences; GSM8K experiments, extended setup details, and noise-robustness results are in Appendix~\ref{app:full_exp} (GSM8K results in Section~\ref{subsec:gsm8k_exp}).

\paragraph{Experimental Setup} We introduce Atom-Task to isolate compositional reasoning, requiring sequential execution of atomic operations on symbolic sequences over a six-symbol alphabet ($\{\alpha, \beta, \gamma, \delta, \varepsilon, \omega\}$). We define five operations: Reverse, Substitute, Duplicate, Deduplicate, and Rotate.

\paragraph{Models and Datasets} For the Atom-Task, we use Qwen3-4B. We design four training datasets: (1) Single-Step: 5k uniformly distributed one-step samples; (2) Three-Step: 1.66k three-step samples; (3) Mixed 3+5: 1,333 samples (833 three-step and 500 five-step; the latter use $\le2$ distinct operations, bridging length and combinatorial generalization); (4) Mixed 1-, 3-, and 5-Step: 4,000 one-step, 600 three-step, and 400 five-step samples (for each operation $op$, a dedicated 5k-sample dataset excludes $op$ from all multi-step chains but keeps it in single-step data). For dataset-size accounting, an $L$-step trajectory contributes $L$ step-level samples; this convention is used only to control the training-data budget. All models are evaluated on held-out 3-, 5-, 7-, and 9-step test sets (500 samples each). Full implementation details are given in Appendix~\ref{app:full_exp}.

\label{subsubsec:aton_results}
\begin{table}[ht!]
\centering
\caption{Accuracy (\%) by training data and test steps. All entries are means over 3 seeds (42, 2026, 12345); standard deviations are shown as subscripts.}
\label{tab:main_results_combined}
\setlength{\tabcolsep}{2.5pt}
\footnotesize
\begin{adjustbox}{max width=\linewidth}
\begin{tabular}{@{}lcccc|lcccc@{}}
\toprule
\multicolumn{5}{c}{\textbf{(a) CoT Setting}} & \multicolumn{5}{c}{\textbf{(b) No-CoT Setting}} \\
\cmidrule(r){1-5} \cmidrule(l){6-10}
\textbf{Data} & \textbf{3-step} & \textbf{5-step} & \textbf{7-step} & \textbf{9-step} & \textbf{Data} & \textbf{3-step} & \textbf{5-step} & \textbf{7-step} & \textbf{9-step} \\
\midrule
1-step & $91.80_{\mbox{\tiny$\pm 2.24$}}$ & $63.13_{\mbox{\tiny$\pm 4.15$}}$ & $36.27_{\mbox{\tiny$\pm 7.87$}}$ & $22.27_{\mbox{\tiny$\pm 3.67$}}$ & 1-step & $10.93_{\mbox{\tiny$\pm 0.84$}}$ & $7.00_{\mbox{\tiny$\pm 0.00$}}$ & $2.73_{\mbox{\tiny$\pm 0.25$}}$ & $0.67_{\mbox{\tiny$\pm 0.34$}}$ \\
3-step & $99.33_{\mbox{\tiny$\pm 0.25$}}$ & $95.73_{\mbox{\tiny$\pm 0.68$}}$ & $88.87_{\mbox{\tiny$\pm 1.54$}}$ & $79.27_{\mbox{\tiny$\pm 3.03$}}$ & 3-step & $75.87_{\mbox{\tiny$\pm 0.84$}}$ & $30.20_{\mbox{\tiny$\pm 0.65$}}$ & $11.00_{\mbox{\tiny$\pm 0.85$}}$ & $5.13_{\mbox{\tiny$\pm 0.09$}}$ \\
Mix$_{3+5}$  & $100.00_{\mbox{\tiny$\pm 0.00$}}$ & $97.87_{\mbox{\tiny$\pm 0.68$}}$ & $94.00_{\mbox{\tiny$\pm 0.59$}}$ & $90.40_{\mbox{\tiny$\pm 1.57$}}$ & Mix$_{3+5}$ & $51.00_{\mbox{\tiny$\pm 0.75$}}$ & $26.60_{\mbox{\tiny$\pm 1.40$}}$ & $12.20_{\mbox{\tiny$\pm 0.75$}}$ & $5.73_{\mbox{\tiny$\pm 0.09$}}$ \\
\bottomrule
\end{tabular}

\end{adjustbox}
\end{table}
\subsection{Main Results and Analysis}

\paragraph{Short-to-Long Generalization \& CoT vs. No-CoT.}
Table~\ref{tab:main_results_combined} compares CoT and No-CoT settings. While CoT with single-step data shows some compositionality, its degradation at longer test steps indicates the benefit of short-chain training data. Crucially, models trained on short chains (3-step/3+5 mixed) perform well on longer tasks; since the 5-step chains limit operation types, this implies models generalize stepwise decomposition patterns. In contrast, No-CoT degradation indicates a failure to capture compositional structures without intermediate steps, causing overfitting to spurious correlations.

\paragraph{Generalization to Unseen Combinations.}
Table~\ref{tab:ablation_combined}(a) shows performance when specific operations are excluded from multi-step samples. Even with 9-step test sequences, all experimental groups maintain high accuracy, indicating that models can effectively compose and generalize operations encountered only in single-step data to multi-step tasks, thereby achieving compositional generalization. Table~\ref{tab:ablation_combined}(b) again confirms the necessity of CoT: without it, performance remains consistently poor regardless of the excluded operation.

\begin{table}[ht!]
\centering
\caption{Atom-Task: Accuracy (\%) when an operation is excluded from multi-step samples. Entries are means over 3 seeds (42, 2026, 12345); standard deviations are shown as subscripts. Rev: Reverse, Dedup: Deduplicate, Sub: Substitute, Dup: Duplicate, Rot: Rotate.}
\label{tab:ablation_combined}
\setlength{\tabcolsep}{2.5pt}
\footnotesize
\begin{adjustbox}{max width=\linewidth}
\begin{tabular}{@{}lcccc|lcccc@{}}
\toprule
\multicolumn{5}{c}{\textbf{(a) CoT Setting}} & \multicolumn{5}{c}{\textbf{(b) No-CoT Setting}} \\
\cmidrule(r){1-5} \cmidrule(l){6-10}
\textbf{Op} & \textbf{3-step} & \textbf{5-step} & \textbf{7-step} & \textbf{9-step} & \textbf{Op} & \textbf{3-step} & \textbf{5-step} & \textbf{7-step} & \textbf{9-step} \\
\midrule
Rev      & $99.53_{\mbox{\tiny$\pm 0.34$}}$ & $97.13_{\mbox{\tiny$\pm 0.25$}}$ & $92.67_{\mbox{\tiny$\pm 1.60$}}$ & $87.13_{\mbox{\tiny$\pm 0.84$}}$ & Rev       & $53.73_{\mbox{\tiny$\pm 0.50$}}$ & $30.33_{\mbox{\tiny$\pm 0.74$}}$ & $16.00_{\mbox{\tiny$\pm 1.18$}}$ & $5.53_{\mbox{\tiny$\pm 0.47$}}$ \\
Dedup   & $97.87_{\mbox{\tiny$\pm 0.47$}}$ & $93.27_{\mbox{\tiny$\pm 1.11$}}$ & $86.20_{\mbox{\tiny$\pm 1.82$}}$ & $76.73_{\mbox{\tiny$\pm 0.66$}}$ & Dedup   & $40.40_{\mbox{\tiny$\pm 0.65$}}$ & $16.93_{\mbox{\tiny$\pm 0.66$}}$ & $6.80_{\mbox{\tiny$\pm 0.43$}}$ & $1.53_{\mbox{\tiny$\pm 0.09$}}$ \\
Sub    & $96.80_{\mbox{\tiny$\pm 0.43$}}$ & $90.73_{\mbox{\tiny$\pm 0.41$}}$ & $84.33_{\mbox{\tiny$\pm 0.38$}}$ & $70.67_{\mbox{\tiny$\pm 0.81$}}$ & Sub   & $42.87_{\mbox{\tiny$\pm 0.81$}}$ & $21.40_{\mbox{\tiny$\pm 0.65$}}$ & $11.20_{\mbox{\tiny$\pm 0.33$}}$ & $5.07_{\mbox{\tiny$\pm 0.50$}}$ \\
Dup    & $93.60_{\mbox{\tiny$\pm 0.59$}}$ & $84.80_{\mbox{\tiny$\pm 0.99$}}$ & $76.73_{\mbox{\tiny$\pm 1.75$}}$ & $66.73_{\mbox{\tiny$\pm 1.25$}}$ & Dup     & $38.87_{\mbox{\tiny$\pm 1.23$}}$ & $17.20_{\mbox{\tiny$\pm 0.75$}}$ & $11.07_{\mbox{\tiny$\pm 1.09$}}$ & $3.40_{\mbox{\tiny$\pm 0.57$}}$ \\
Rot        & $96.53_{\mbox{\tiny$\pm 0.66$}}$ & $91.93_{\mbox{\tiny$\pm 0.66$}}$ & $86.93_{\mbox{\tiny$\pm 0.34$}}$ & $80.13_{\mbox{\tiny$\pm 0.41$}}$ & Rot       & $55.47_{\mbox{\tiny$\pm 0.41$}}$ & $35.60_{\mbox{\tiny$\pm 1.84$}}$ & $22.67_{\mbox{\tiny$\pm 1.09$}}$ & $9.73_{\mbox{\tiny$\pm 0.93$}}$ \\
\bottomrule
\end{tabular}

\end{adjustbox}
\end{table}

\begin{figure}[htbp]
\centering
\includegraphics[width=\linewidth]{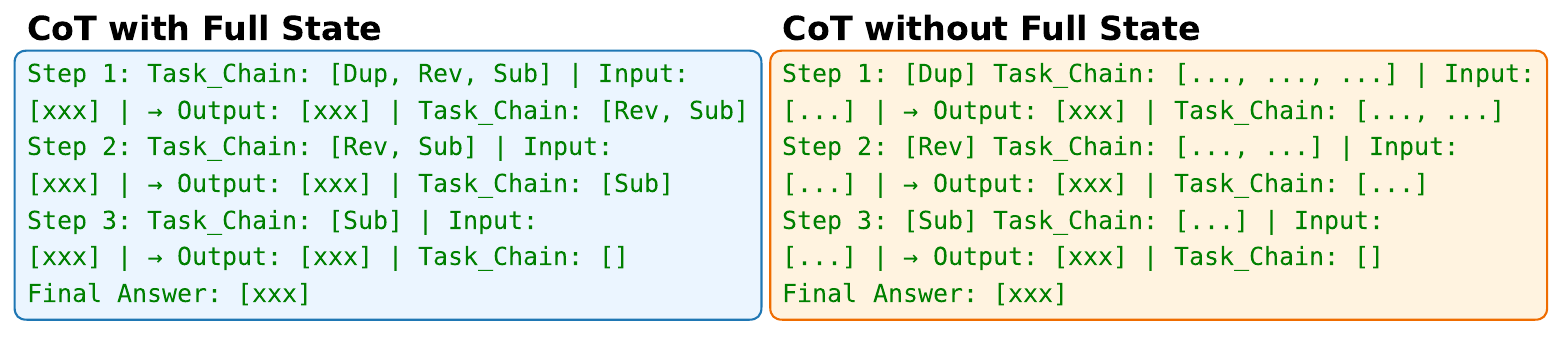}
\caption{Comparison of complete vs. incomplete CoT state recordings.}
\label{fig:fig2}
\end{figure}

\paragraph{Effect of State Tracking in CoT.}
We ablate the CoT format by removing explicit state tracking (remaining tasks and input context) from the intermediate outputs (shown in Figure~\ref{fig:fig2}), such that the intermediate state $z_{i-1}$ can no longer carry complete information. Table~\ref{tab:format_ablation_merged} shows this ablation causes a consistent performance degradation (see the drop in average accuracy on the 7-step test set). This demonstrates that when $z_{i-1}$ fails to preserve the full state, multi-step decomposition is impaired. This validates our theoretical requirement that $z_{i-1}$ must carry complete state information.

\begin{table}[ht!]
\centering
\caption{Format ablation: Accuracy (\%) with vs. without explicit state tracking. Abl.: means over 3 seeds (42, 2026, 12345); Orig.: means over 3 seeds from Table~\ref{tab:ablation_combined}.}
\label{tab:format_ablation_merged}
\footnotesize
\begin{adjustbox}{max width=\linewidth}
\begin{tabular}{@{}lccccc|cc@{}}
\toprule
\textbf{Step} & \textbf{Reverse} & \textbf{Rotate} & \textbf{Duplicate} & \textbf{Deduplicate} & \textbf{Substitute} & \textbf{Avg. (Abl.)} & \textbf{Avg. (Orig.)} \\
\midrule
3-step & $98.47_{\mbox{\tiny$\pm 1.05$}}$ & $95.53_{\mbox{\tiny$\pm 2.08$}}$ & $92.33_{\mbox{\tiny$\pm 4.97$}}$ & $96.93_{\mbox{\tiny$\pm 1.43$}}$ & $94.53_{\mbox{\tiny$\pm 4.20$}}$ & $95.56_{\mbox{\tiny$\pm 1.41$}}$ & $96.87_{\mbox{\tiny$\pm 0.23$}}$ \\
5-step & $95.80_{\mbox{\tiny$\pm 2.69$}}$ & $88.13_{\mbox{\tiny$\pm 5.12$}}$ & $83.20_{\mbox{\tiny$\pm 6.95$}}$ & $91.73_{\mbox{\tiny$\pm 5.49$}}$ & $87.40_{\mbox{\tiny$\pm 7.08$}}$ & $89.25_{\mbox{\tiny$\pm 2.55$}}$ & $91.57_{\mbox{\tiny$\pm 0.34$}}$ \\
7-step & $79.40_{\mbox{\tiny$\pm 10.16$}}$ & $67.80_{\mbox{\tiny$\pm 2.61$}}$ & $67.47_{\mbox{\tiny$\pm 6.13$}}$ & $62.47_{\mbox{\tiny$\pm 4.43$}}$ & $60.67_{\mbox{\tiny$\pm 20.02$}}$ & $67.56_{\mbox{\tiny$\pm 4.77$}}$ & $85.37_{\mbox{\tiny$\pm 0.61$}}$ \\
\bottomrule
\end{tabular}
\end{adjustbox}
\end{table}

\begin{figure}[ht!]
    \centering
    \includegraphics[width=\linewidth]{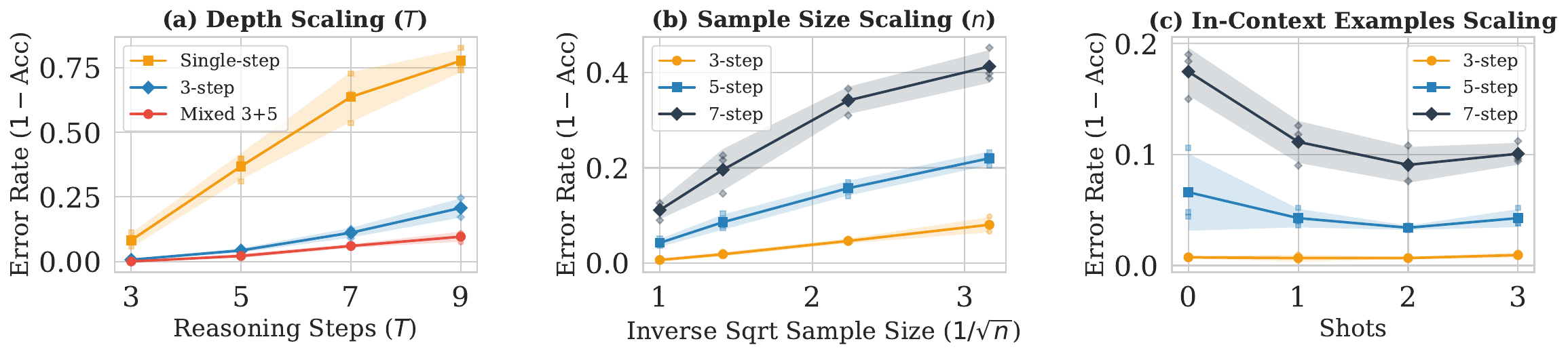}
    \caption{Each entry is a mean over 3 seeds (42, 2026, 12345). Atom-Task results when varying (a) test depth, (b) training sample size, and (c) inference-time examples; in (b) the horizontal axis is $1/\sqrt{n}$, where $n$ is the sample size in units of the full budget ($10\%$, $20\%$, $50\%$, $100\%$).}
    \label{fig:scaling}
\end{figure}

\paragraph{Scaling Trends of Compositional Generalization}
\label{subsubsec:scaling}
We study performance as a function of reasoning depth, training sample size, and the number of inference-time examples, as summarized in Figure~\ref{fig:scaling}. 
For depth scaling (Figure~\ref{fig:scaling}a), error increases with $T$, with diminishing increments for single-step training. Theorem~\ref{lem:rad} gives sublinear-in-$T$ control of the estimation term at fixed $n$.
For sample scaling (Figure~\ref{fig:scaling}b), increasing the sample size consistently reduces the test error rate, particularly for longer reasoning chains, which is qualitatively consistent with a smaller estimation error at larger sample sizes. Figure~\ref{fig:scaling}c varies the number of in-context examples at inference time rather than the theoretical sliding-window length $m$. Detailed data for these scaling experiments are provided in Appendix~\ref{subsubsec:scaling_details}.

\section{Further Related Work}
\label{sec:related_work}

\paragraph{CoT Prompting and Training}
Chain-of-Thought (CoT) enhances complex problem-solving by generating intermediate reasoning steps \cite{wei2022chain}, improving performance on math benchmarks \cite{cobbe2021training}. This led to extensions such as Tree of Thoughts \cite{yao2023tree}, Graph of Thoughts \cite{besta2024graph}, and Program of Thoughts \cite{chenprogram}, alongside robustness improvements via self-consistency \cite{wangself}, self-refinement \cite{madaan2023self}, and retrieval-augmented generation \cite{lewis2020retrieval}. Decomposition methods like Least-to-Most prompting \cite{zhouleast} and Automatic CoT \cite{zhangautomatic} further automate this process. However, most work focuses on inference-stage prompts, with limited discussion of CoT's intrinsic training mechanisms \cite{prystawski2023think}. Training LLMs with CoT via supervised fine-tuning or reinforcement learning, on the other hand, enhances capability \cite{yumetamath,shao2024deepseekmath,zelikman2022star} and improves compositional generalization \cite{yao2026compositional}.

\paragraph{CoT Theoretical Analysis}
Theoretically, CoT increases Transformers' effective depth for serial problems \cite{merrillexpressive,li2024chain}, and linear next-token predictors on CoT data can approximate Turing-computable functions \cite{feng2023towards}. Further analyses cover sample/time complexity \cite{joshi2025theory}, task decomposition \cite{yang2025chain}, in-context learning bounds \cite{bavandpour2025lower}, and expressive power on arithmetic or gradient descent \cite{huangtransformers}, with structural analysis showing reasoning internalized as combinatorial circuits \cite{wang2024grokking,biran2024hopping}. However, these studies are often qualitative or limited to localized tasks \cite{gandhicognitive,prystawski2023think}. More specifically, Joshi et al.~\cite{joshi2025theory} and Hanneke et al.~\cite{hanneke2026sample} analyze the PAC (probably approximately correct) error of the final symbol under complete-trajectory supervision and realizability, whereas our setting supervises individual atomic steps through a target-token objective and provides explicit depth-dependent bounds. The analysis of \cite{yao2026compositional} considers out-of-distribution generalization but is restricted to two-step tasks and lacks LLM-specific analysis. Maurer et al.~\cite{maurer2016benefit} provide a shared-representation perspective for pooled estimation, whereas Li et al.~\cite{li2025generalization} provide the decoder covering-number tool used for our restricted final-$k$ output class.

\section{Conclusion and Future Work}
\label{sec:conclusion}
In this work we asked where the out-of-distribution capability of CoT comes from, and located it in the reuse of one transition model along the chain. The analysis separates a multiplicative price for inference errors, of order $\mathcal{O}(k)$, set by the block length rather than by the number of steps, from a statistical cost that grows only as $\mathcal{O}(\sqrt{T/n})$, since a single predictor serves all $T$ sub-tasks and can be assessed on a combined sample. In the token-by-token prediction regime this gives an end-to-end bound that is $B_\ell k/\log 2$ times the best-in-class summed step-level risk, plus an estimation term of order $\widetilde{\mathcal O}(k\sqrt{T/n})$; a matched construction for the sub-task ERM shows that the $\sqrt{T/n}$ dependence is not an artefact of the analysis. The decomposition further indicates how learning reusable transitions can support compositional transfer---consistent with our experiments, where explicit state tracking proves essential. Taken together, the analysis sharpens our picture of how compositional OOD generalization arises under CoT, and that picture is not tied to autoregressive token-by-token output: it may also inform future analyses of agents that maintain and update an external memory.

\paragraph{Limitations \& Future Work}
Our theory studies explicit CoT under supervised NTP-style training, where intermediate states and step-level transitions are observable; it excludes adaptive paradigms such as reinforcement fine-tuning and test-time search \cite{swamy2025all,setlur2025scaling,yuedoes,gandhicognitive,shah2025rethinking}, where trajectories may be selected, refined, or rewarded. Extending our decomposition framework to these settings is a natural future direction, and does not affect the mechanism identified here.

\acksection

This work was supported by the Special Fund for Key Program of Science and Technology of
Jiangsu Province (No.~BG2024042); the Gusu Leading Talents Program for Innovation and
Entrepreneurship (No.~ZXL2025323); the Fundamental and Interdisciplinary Disciplines
Breakthrough Plan of the Ministry of Education of China (No.~JYB2025XDXM118); and the
``111 Center'' (No.~B26023). This work was also partially supported by the Lee Kong Chian
Fellowship awarded to Pan Zhou by Singapore Management University. We thank the NeurIPS
reviewers for their valuable feedback and suggestions.
\bibliographystyle{unsrt}
\bibliography{references}

@article{wei2022chain,
  title={Chain-of-thought prompting elicits reasoning in large language models},
  author={Wei, Jason and Wang, Xuezhi and Schuurmans, Dale and Bosma, Maarten and Xia, Fei and Chi, Ed and Le, Quoc V and Zhou, Denny and others},
  journal={Advances in neural information processing systems},
  volume={35},
  pages={24824--24837},
  year={2022}
}

@inproceedings{merrillexpressive,
  title={The Expressive Power of Transformers with Chain of Thought},
  author={Merrill, William and Sabharwal, Ashish},
  booktitle={The Twelfth International Conference on Learning Representations},
  year={2024}
}

@inproceedings{li2024chain,
  title={Chain of thought empowers transformers to solve inherently serial problems},
  author={Li, Zhiyuan and Liu, Hong and Zhou, Denny and Ma, Tengyu},
  booktitle={The Twelfth International Conference on Learning Representations},
  year={2024}
}

@inproceedings{joshi2025theory,
  title={A Theory of Learning with Autoregressive Chain of Thought},
  author={Joshi, Nirmit and Vardi, Gal and Block, Adam and Goel, Surbhi and Li, Zhiyuan and Misiakiewicz, Theodor and Srebro, Nathan},
  booktitle={The Thirty Eighth Annual Conference on Learning Theory},
  pages={3161--3212},
  year={2025},
  organization={PMLR}
}

@inproceedings{yao2026compositional,
  title={Compositional Generalization from Learned Skills via CoT Training: A Theoretical and Structural Analysis for Reasoning},
  author={Yao, Xinhao and Ren, Ruifeng and Liao, Yun and Ding, Lizhong and Liu, Yong},
  booktitle={The Fourteenth International Conference on Learning Representations},
  year={2026}
}

@inproceedings{yang2025chain,
  title={Chain-of-thought provably enables learning the (otherwise) unlearnable},
  author={Yang, Chenxiao and Li, Zhiyuan and Wipf, David},
  booktitle={The Thirteenth International Conference on Learning Representations},
  year={2025}
}

@article{yao2023tree,
  title={Tree of thoughts: Deliberate problem solving with large language models},
  author={Yao, Shunyu and Yu, Dian and Zhao, Jeffrey and Shafran, Izhak and Griffiths, Tom and Cao, Yuan and Narasimhan, Karthik},
  journal={Advances in neural information processing systems},
  volume={36},
  pages={11809--11822},
  year={2023}
}

@inproceedings{besta2024graph,
  title={Graph of thoughts: Solving elaborate problems with large language models},
  author={Besta, Maciej and Blach, Nils and Kubicek, Ales and Gerstenberger, Robert and Podstawski, Michal and Gianinazzi, Lukas and Gajda, Joanna and Lehmann, Tomasz and Niewiadomski, Hubert and Nyczyk, Piotr and others},
  booktitle={Proceedings of the AAAI conference on artificial intelligence},
  volume={38},
  number={16},
  pages={17682--17690},
  year={2024}
}

@article{chenprogram,
  title={Program of Thoughts Prompting: Disentangling Computation from Reasoning for Numerical Reasoning Tasks},
  author={Chen, Wenhu and Ma, Xueguang and Wang, Xinyi and Cohen, William W},
  journal={Transactions on Machine Learning Research},
  year={2023}
}

@inproceedings{zhangautomatic,
  title={Automatic Chain of Thought Prompting in Large Language Models},
  author={Zhang, Zhuosheng and Zhang, Aston and Li, Mu and Smola, Alex},
  booktitle={The Eleventh International Conference on Learning Representations},
  year={2023}
}

@inproceedings{zhouleast,
  title={Least-to-Most Prompting Enables Complex Reasoning in Large Language Models},
  author={Zhou, Denny and Sch{\"a}rli, Nathanael and Hou, Le and Wei, Jason and Scales, Nathan and Wang, Xuezhi and Schuurmans, Dale and Cui, Claire and Bousquet, Olivier and Le, Quoc V and others},
  booktitle={The Eleventh International Conference on Learning Representations},
  year={2023}
}

@article{madaan2023self,
  title={Self-refine: Iterative refinement with self-feedback},
  author={Madaan, Aman and Tandon, Niket and Gupta, Prakhar and Hallinan, Skyler and Gao, Luyu and Wiegreffe, Sarah and Alon, Uri and Dziri, Nouha and Prabhumoye, Shrimai and Yang, Yiming and others},
  journal={Advances in neural information processing systems},
  volume={36},
  pages={46534--46594},
  year={2023}
}

@article{lewis2020retrieval,
  title={Retrieval-augmented generation for knowledge-intensive nlp tasks},
  author={Lewis, Patrick and Perez, Ethan and Piktus, Aleksandra and Petroni, Fabio and Karpukhin, Vladimir and Goyal, Naman and K{\"u}ttler, Heinrich and Lewis, Mike and Yih, Wen-tau and Rockt{\"a}schel, Tim and others},
  journal={Advances in neural information processing systems},
  volume={33},
  pages={9459--9474},
  year={2020}
}

@inproceedings{li2025generalization,
  title={On the generalization ability of next-token-prediction pretraining},
  author={Li, Zhihao and Jiang, Xue and Liu, Liyuan and Zhang, Xuelin and Chen, Hong and Zheng, Feng},
  booktitle={Forty-second International Conference on Machine Learning},
  year={2025}
}

@inproceedings{bavandpour2025lower,
  title={Lower Bounds for Chain-of-Thought Reasoning in Hard-Attention Transformers},
  author={Bavandpour, Alireza Amiri and Huang, Xinting and Rofin, Mark and Hahn, Michael},
  booktitle={International Conference on Machine Learning},
  pages={3274--3306},
  year={2025},
  organization={PMLR}
}

@article{feng2023towards,
  title={Towards revealing the mystery behind chain of thought: a theoretical perspective},
  author={Feng, Guhao and Zhang, Bohang and Gu, Yuntian and Ye, Haotian and He, Di and Wang, Liwei},
  journal={Advances in Neural Information Processing Systems},
  volume={36},
  pages={70757--70798},
  year={2023}
}

@inproceedings{gandhicognitive,
  title={Cognitive Behaviors that Enable Self-Improving Reasoners, or, Four Habits of Highly Effective STaRs},
  author={Gandhi, Kanishk and Chakravarthy, Ayush K and Singh, Anikait and Lile, Nathan and Goodman, Noah},
  booktitle={Second Conference on Language Modeling},
  year={2025}
}

@article{guo2025deepseek,
  title={Deepseek-r1: Incentivizing reasoning capability in llms via reinforcement learning},
  author={Guo, Daya and Yang, Dejian and Zhang, Haowei and Song, Junxiao and Wang, Peiyi and Zhu, Qihao and Xu, Runxin and Zhang, Ruoyu and Ma, Shirong and Bi, Xiao and others},
  journal={arXiv preprint arXiv:2501.12948},
  year={2025}
}

@inproceedings{wangself,
  title={Self-Consistency Improves Chain of Thought Reasoning in Language Models},
  author={Wang, Xuezhi and Wei, Jason and Schuurmans, Dale and Le, Quoc V and Chi, Ed H and Narang, Sharan and Chowdhery, Aakanksha and Zhou, Denny},
  booktitle={The Eleventh International Conference on Learning Representations},
  year={2023}
}

@article{cobbe2021training,
  title={Training verifiers to solve math word problems},
  author={Cobbe, Karl and Kosaraju, Vineet and Bavarian, Mohammad and Chen, Mark and Jun, Heewoo and Kaiser, Lukasz and Plappert, Matthias and Tworek, Jerry and Hilton, Jacob and Nakano, Reiichiro and others},
  journal={arXiv preprint arXiv:2110.14168},
  year={2021}
}

@inproceedings{yumetamath,
  title={MetaMath: Bootstrap Your Own Mathematical Questions for Large Language Models},
  author={Yu, Longhui and Jiang, Weisen and Shi, Han and YU, Jincheng and Liu, Zhengying and Zhang, Yu and Kwok, James and Li, Zhenguo and Weller, Adrian and Liu, Weiyang},
  booktitle={The Twelfth International Conference on Learning Representations},
  year={2024}
}

@article{shao2024deepseekmath,
  title={Deepseekmath: Pushing the limits of mathematical reasoning in open language models},
  author={Shao, Zhihong and Wang, Peiyi and Zhu, Qihao and Xu, Runxin and Song, Junxiao and Bi, Xiao and Zhang, Haowei and Zhang, Mingchuan and Li, YK and Wu, Yang and others},
  journal={arXiv preprint arXiv:2402.03300},
  year={2024}
}

@article{zelikman2022star,
  title={Star: Bootstrapping reasoning with reasoning},
  author={Zelikman, Eric and Wu, Yuhuai and Mu, Jesse and Goodman, Noah},
  journal={Advances in Neural Information Processing Systems},
  volume={35},
  pages={15476--15488},
  year={2022}
}

@article{prystawski2023think,
  title={Why think step by step? reasoning emerges from the locality of experience},
  author={Prystawski, Ben and Li, Michael and Goodman, Noah},
  journal={Advances in Neural Information Processing Systems},
  volume={36},
  pages={70926--70947},
  year={2023}
}

@article{wang2024grokking,
  title={Grokking of implicit reasoning in transformers: A mechanistic journey to the edge of generalization},
  author={Wang, Boshi and Yue, Xiang and Su, Yu and Sun, Huan},
  journal={Advances in Neural Information Processing Systems},
  volume={37},
  pages={95238--95265},
  year={2024}
}

@inproceedings{biran2024hopping,
  title={Hopping too late: Exploring the limitations of large language models on multi-hop queries},
  author={Biran, Eden and Gottesman, Daniela and Yang, Sohee and Geva, Mor and Globerson, Amir},
  booktitle={Proceedings of the 2024 Conference on Empirical Methods in Natural Language Processing},
  pages={14113--14130},
  year={2024}
}

@inproceedings{huangtransformers,
  title={Transformers Learn to Implement Multi-step Gradient Descent with Chain of Thought},
  author={Huang, Jianhao and Wang, Zixuan and Lee, Jason D},
  booktitle={The Thirteenth International Conference on Learning Representations},
  year={2025}
}

@article{swamy2025all,
  title={All roads lead to likelihood: The value of reinforcement learning in fine-tuning},
  author={Swamy, Gokul and Choudhury, Sanjiban and Sun, Wen and Wu, Zhiwei Steven and Bagnell, J Andrew},
  journal={arXiv preprint arXiv:2503.01067},
  year={2025}
}

@inproceedings{setlur2025scaling,
  title={Scaling Test-Time Compute Without Verification or RL is Suboptimal},
  author={Setlur, Amrith and Rajaraman, Nived and Levine, Sergey and Kumar, Aviral},
  booktitle={International Conference on Machine Learning},
  pages={54058--54094},
  year={2025},
  organization={PMLR}
}

@article{shah2025rethinking,
  title={Rethinking reflection in pre-training},
  author={Shah, Darsh J and Rushton, Peter and Singla, Somanshu and Parmar, Mohit and Smith, Kurt and Vanjani, Yash and Vaswani, Ashish and Chaluvaraju, Adarsh and Hojel, Andrew and Ma, Andrew and others},
  journal={arXiv preprint arXiv:2504.04022},
  year={2025}
}

@inproceedings{yuedoes,
  title={Does Reinforcement Learning Really Incentivize Reasoning Capacity in LLMs Beyond the Base Model?},
  author={Yue, Yang and Chen, Zhiqi and Lu, Rui and Zhao, Andrew and Wang, Zhaokai and Song, Shiji and Huang, Gao},
  booktitle={The Thirty-ninth Annual Conference on Neural Information Processing Systems},
  year={2025}
}

@inproceedings{kim2023cot,
  title={The cot collection: Improving zero-shot and few-shot learning of language models via chain-of-thought fine-tuning},
  author={Kim, Seungone and Joo, Se and Kim, Doyoung and Jang, Joel and Ye, Seonghyeon and Shin, Jamin and Seo, Minjoon},
  booktitle={Proceedings of the 2023 Conference on Empirical Methods in Natural Language Processing},
  pages={12685--12708},
  year={2023}
}

@inproceedings{prabhakar2024deciphering,
  title={Deciphering the factors influencing the efficacy of chain-of-thought: Probability, memorization, and noisy reasoning},
  author={Prabhakar, Akshara and Griffiths, Thomas L and McCoy, R Thomas},
  booktitle={Findings of the Association for Computational Linguistics: EMNLP 2024},
  pages={3710--3724},
  year={2024}
}

@inproceedings{yuemammoth,
  title={MAmmoTH: Building Math Generalist Models through Hybrid Instruction Tuning},
  author={Yue, Xiang and Qu, Xingwei and Zhang, Ge and Fu, Yao and Huang, Wenhao and Sun, Huan and Su, Yu and Chen, Wenhu},
  booktitle={The Twelfth International Conference on Learning Representations},
  year={2024}
}

@inproceedings{wang2024math,
  title={Math-shepherd: Verify and reinforce llms step-by-step without human annotations},
  author={Wang, Peiyi and Li, Lei and Shao, Zhihong and Xu, Runxin and Dai, Damai and Li, Yifei and Chen, Deli and Wu, Yu and Sui, Zhifang},
  booktitle={Proceedings of the 62nd Annual Meeting of the Association for Computational Linguistics (Volume 1: Long Papers)},
  pages={9426--9439},
  year={2024}
}

@article{yu2024flow,
  title={Flow of reasoning: Efficient training of llm policy with divergent thinking},
  author={Yu, Fangxu},
  journal={arXiv (Cornell University)},
  year={2024}
}

@inproceedings{ho2023large,
  title={Large language models are reasoning teachers},
  author={Ho, Namgyu and Schmid, Laura and Yun, Se-Young},
  booktitle={Proceedings of the 61st annual meeting of the association for computational linguistics (volume 1: long papers)},
  pages={14852--14882},
  year={2023}
}

@article{openai2024learning,
  title={Learning to reason with LLMs},
  author={OpenAI, Team},
  journal={OpenAI Blog},
  year={2024}
}

@book{shalev2014understanding,
  title={Understanding machine learning: From theory to algorithms},
  author={Shalev-Shwartz, Shai and Ben-David, Shai},
  year={2014},
  publisher={Cambridge university press}
}

@article{liu2022gradient,
  title={Gradient learning under tilted empirical risk minimization},
  author={Liu, Liyuan and Song, Biqin and Pan, Zhibin and Yang, Chuanwu and Xiao, Chi and Li, Weifu},
  journal={Entropy},
  volume={24},
  number={7},
  pages={956},
  year={2022},
  publisher={MDPI}
}

@inproceedings{deng2024generalization,
  title={On the generalization ability of unsupervised pretraining},
  author={Deng, Yuyang and Hong, Junyuan and Zhou, Jiayu and Mahdavi, Mehrdad},
  booktitle={International Conference on Artificial Intelligence and Statistics},
  pages={4519--4527},
  year={2024},
  organization={PMLR}
}

@article{liu2025consistency,
  title={The consistency analysis of gradient learning under independent covariate shift},
  author={Liu, Liyuan and Chen, Hong and Xiao, Chi and Li, Weifu},
  journal={Neurocomputing},
  volume={635},
  pages={129883},
  year={2025},
  publisher={Elsevier}
}

@article{abedsoltan2025task,
  title={Task Generalization With AutoRegressive Compositional Structure: Can Learning From $D$ Tasks Generalize to $D^{T}$ Tasks?},
  author={Abedsoltan, Amirhesam and Zhang, Huaqing and Wen, Kaiyue and Lin, Hongzhou and Zhang, Jingzhao and Belkin, Mikhail},
  journal={arXiv preprint arXiv:2502.08991},
  year={2025}
}

@article{maurer2016benefit,
  title={The benefit of multitask representation learning},
  author={Maurer, Andreas and Pontil, Massimiliano and Romera-Paredes, Bernardino},
  journal={Journal of Machine Learning Research},
  volume={17},
  number={81},
  pages={1--32},
  year={2016}
}

@article{hanneke2026sample,
  title   = {Sample Complexity of Autoregressive Reasoning: Chain-of-Thought vs. End-to-End},
  author  = {Hanneke, Steve and Mehalel, Idan and Moran, Shay},
  journal = {arXiv preprint arXiv:2604.12013},
  year    = {2026}
}

\clearpage
\appendix
\section{Supplementary Details on NTP Setting}
\label{app:ntp_setting}

This section supplements Section~\ref{sec:ntp} by defining the decoder architecture and the sampling units used in the target-token analysis. We use the convention of Section~\ref{sec:ntp_form}: $m$ denotes the complete-record length (and the maximum sequence length used by the decoder), the block-boundary state has length $m-k$, and $\mathcal{V}$ denotes the vocabulary.

\subsection{Decoder-only Model Architecture}
\label{app:dec_arch}

Contemporary Large Language Models (LLMs) predominantly utilize a decoder-only Transformer architecture. As described in Section~\ref{sec:ntp_form}, the composite model $\mathrm{LM} = g \circ h$ consists of a representation learner $h$ and a token predictor $g$. We formalize the representation learner $h$ as an $L$-layer Transformer decoder, following the block description of \cite[Eq.~(4)]{li2025generalization}, mapping an input token sequence $Z \in \mathcal{V}^m$ to a hidden representation $h(Z) \in \mathbb{R}^{m \times d}$. The decoder applies to any prefix $Z \in \mathcal{V}^s$ with $s\le m$, and the prediction head reads its last hidden state.

The process begins with an Embedding Layer, which converts discrete tokens into continuous vectors $Z_0 \in \mathbb{R}^{m \times d}$ via:
\begin{equation*}
    Z_0 = \text{Embedding}(Z) := \mathrm{OneHot}(Z) W_e + W_p,
\end{equation*}
where $\mathrm{OneHot}(Z)\in\{0,1\}^{m\times n_v}$ is the one-hot encoding of the $m$ tokens, $W_e \in \mathbb{R}^{n_v \times d}$ is the token embedding matrix, and $W_p \in \mathbb{R}^{m \times d}$ is the position embedding matrix. The core of the architecture consists of $L$ stacked Transformer-decoder blocks. For layer $l \in \{1, \dots, L\}$, the output $Z_l$ is computed from $Z_{l-1}$ via masked self-attention and feed-forward networks (FFN). Specifically, let $H$ be the number of attention heads and $d_k = d/H$. The masked self-attention head $h \in [H]$ is defined as:
\begin{equation*}
    \begin{gathered}
\text{Attn}_h(Z_{l-1}) = \text{softmax}\left(\frac{Q_h K_h^\top}{\sqrt{d_k}} + M\right) V_h,\\
Q_h=Z_{l-1}W_h^Q,\quad K_h=Z_{l-1}W_h^K,\quad V_h=Z_{l-1}W_h^V.
\end{gathered}
\end{equation*}
Here, $M \in \mathbb{R}^{m \times m}$ is a causal mask matrix where $M_{ij} = -\infty$ if $j > i$ and $0$ otherwise. This mask ensures that the hidden state at position $t$ attends only to tokens at positions $\le t$, so that the token at position $t$ is predicted from the state at position $t-1$, thereby implementing the autoregressive property essential for Next-Token Prediction (NTP). The final output of the layer is given by:
\begin{equation*}
    Y_l := \text{Norm}_1\left(\sum_{h=1}^H \text{Attn}_h(Z_{l-1})W_h^O + Z_{l-1}\right), \quad Z_l = \text{Norm}_2\left(\text{FFN}(Y_l) + Y_l\right),
\end{equation*}
where $\text{Norm}_1,\text{Norm}_2$ denote Layer Normalization and $\text{FFN}(Y)=\sigma(YW_{F1})W_{F2}$ is a two-layer feed-forward network with hidden width $d_f=4d$. The representation learner thus serializes the input into a deep representation $h(Z) = Z_L$.

The token predictor $g$ applies a linear projection to the hidden state at the last position, $g(h(Z)) = \text{softmax}((h(Z))_{s,:} W_P)$ with $s=|Z|$ and $W_P\in\mathbb R^{d\times n_v}$, mapping it to a probability distribution over the vocabulary $\mathcal{V}$.

\subsection{Explicit Parameterization and Derivation of the Dimension-Dependent Bound}
\label{app:dim_derivation}

To quantify the Rademacher term in Corollary~\ref{cor:erm_final_risk}, we use the matrix-covering tools recorded in \cite[Lemma~C.3]{li2025generalization}. The statistical argument below is carried out directly for sequence-level losses.

\begin{assumption}[Capacity and Regularity Constraints]
\label{assum:arch}
For the $L$-layer, $H$-head decoder-only Transformer defined in Appendix~\ref{app:dec_arch} (following \cite[Eq.~(4) and Assumption~4.14]{li2025generalization}), we assume:
\begin{itemize}
    \item The Layer Normalization operators $\Pi_{\text{norm}}$ (the operators $\text{Norm}_1$ and $\text{Norm}_2$ of Appendix~\ref{app:dec_arch}) are $G_\pi$-Lipschitz continuous with respect to the 2-norm.
    \item For all layers $l\in[L]$ and attention heads $h\in[H]$, the scaled attention scores satisfy
\[\|Q_l^h(K_l^h)^\top/\sqrt{d_k}\|_{\ell_\infty}\le C_l.\]
    \item Each attention and feed-forward parameter matrix in layer $l$ has Frobenius norm at most $B_l$.
\end{itemize}
The embedding matrices and output head also have bounded Frobenius norms. Let $B_*\ge1$ be a common upper bound for these norms and the $B_l$. For the capacity estimate, normalization is a fixed rowwise operator whose output rows have norm at most $C_\pi\sqrt d$, with fixed $C_\pi$; the feed-forward activation is 1-Lipschitz and vanishes at zero, as does ReLU. These are the usual bounded-row normalization and feed-forward conditions. The constants $B_*,G_\pi,C_\pi$ are fixed independently of the sample size and model dimensions. Let $\Theta$ count all trainable scalar parameters, including token and positional embeddings whenever they are trained, and the output head.

The norm and Lipschitz upper bounds may be enlarged so that $B_l,G_\pi\ge1$. Let $N=nT$ and $\rho_L=\sum_{l=1}^{L}B_l^2$, and set
\begin{align*}
s_L(m,N)
&=\prod_{l=1}^{L}G_\pi^2(B_l^2+1)
\left(e^{C_l}B_l^2H\sqrt N\,m d+1\right),\\
\tau_{m,N}
&=\log\!\left(1+\rho_Ls_L(m,N)\right).
\end{align*}
Here $s_L$ collects the per-layer amplification factors that a parameter perturbation of size $\eta$ accumulates over the $L$ blocks, and $\tau_{m,N}$ is the logarithm of the resulting covering-number bound; the individual factors are derived in the propagation estimate below.
\end{assumption}
For fixed norm, attention-score, and normalization bounds, $\tau_{m,N}=\mathcal O(L\log(2+NHmd)+\log(1+L))$. Below, constants in $\lesssim$ may depend on the fixed bounds $B_*,G_\pi,C_\pi$, but not on $G_\ell,N,m,k,L,d,H,n_v$.

\begin{proof}[Derivation of Eq.~\eqref{eq:qk_capacity}]
We first record a coarse output cover, including every trainable matrix. A Frobenius ball of radius $B_*$ in $\mathbb R^{a\times b}$ has an $\eta$-net, i.e., a set of points whose $\eta$-neighbourhood covers the ball, of cardinality at most $(1+2B_*/\eta)^{ab}$, by vectorization and \cite[Lemma~C.3]{li2025generalization}. Taking the product of these nets gives at most $(1+2B_*/\eta)^\Theta$ parameter representatives. Rather than composing the layerwise coverings of \cite[Lemma~4.15 and Proposition~4.17]{li2025generalization}, we propagate the approximation errors through the actual block of Appendix~\ref{app:dec_arch}, which keeps our normalization placement unchanged.

For completeness, write $\mathrm{TF}_l$ for block $l$. A normalized row has norm at most $C_\pi\sqrt d$, and the bounded feed-forward weights and residual therefore give a block-output row norm at most a constant times $\sqrt d$. Embedding rows satisfy the same type of bound. The causal softmax has Euclidean Lipschitz constant at most $1$, and its attention matrix has operator norm at most $\sqrt m$. For two such admissible inputs, if every matrix in the two blocks differs by at most $\eta$, multiplication of the query, key, value, and output matrices, followed by the residual and feed-forward operations, gives
\begin{equation*}
\|\mathrm{TF}_l(Z)-\mathrm{TF}'_l(Z')\|_F
\le \Gamma\bigl(\|Z-Z'\|_F+\sqrt m\,\eta\bigr),
\qquad \Gamma=cHmd^2\ge1,
\end{equation*}
where $c$ depends only on $B_*,G_\pi,C_\pi$. Indeed, the query-key input perturbation is bounded by a constant times $\sqrt{md}\|Z-Z'\|_F$, its parameter perturbation by a constant times $md\eta$, and the remaining projections contribute at most a constant times $\sqrt{md}$; the displayed $\Gamma$ bounds all these factors. The embedding approximation error is at most $2\sqrt m\,\eta$. Induction over the $L$ blocks and inclusion of the output head thus give, for the concatenated probability outputs on $N$ records,
\begin{equation*}
    \|A-A'\|_F\le\sqrt{Nm}\,\Lambda\eta,
    \qquad \Lambda=(L+3)\Gamma^{L+1}.
\end{equation*}
This argument covers trainable embeddings as part of the parameter vector; fixed embeddings require no net. Covering representatives can be taken in the norm balls, and recentering in the original function class, if necessary, only changes numerical constants.

Let $\Pi_k$ select the final $k$ prediction rows, aligned with the target labels as in Eq.~\eqref{eq:target_token_loss}. Every such row is computed from its teacher-forced prefix. Since
\begin{equation*}
    \|\Pi_kA-\Pi_kB\|_F\le\|A-B\|_F,
\end{equation*}
a full-output cover also covers the selected rows. For one record, write $q(A)$ for its target-token loss with the labels fixed. The loss Lipschitz condition and Cauchy--Schwarz give
\begin{align*}
    |q(A)-q(B)|
    &\le\frac{G_\ell}{k}\sum_{r=1}^{k}
      \|A_{m-k+r}-B_{m-k+r}\|_2\\
    &\le\frac{G_\ell}{\sqrt k}\|\Pi_kA-\Pi_kB\|_F.
\end{align*}
Hence the concatenated $N$-dimensional loss vectors are approximated within $G_\ell\sqrt{Nm/k}\,\Lambda\eta$ (aggregating the $N$ records in $\ell_2$). Choosing
$\eta=\epsilon\sqrt{k}/(G_\ell\sqrt{Nm}\Lambda)$, and setting $\epsilon=G_\ell\sqrt N\,u$, yields the covering-number bound
\begin{equation*}
\log\mathcal N(\mathcal Q_k|_{S_0},G_\ell\sqrt N\,u,\|\cdot\|_2)
\le\Theta\log\left(1+\frac{2B_*\Lambda\sqrt{m/k}}{u}\right),
\end{equation*}

where $\mathcal N(\cdot,\cdot,\|\cdot\|_2)$ denotes the covering number, i.e., the smallest cardinality of such a net in the $\ell_2$ norm, and $\mathcal Q_k|_{S_0}$ is the restriction of $\mathcal Q_k$ to the sample $S_0$. Center the loss vectors at any fixed member of $\mathcal Q_k|_{S_0}$. This leaves their empirical Rademacher complexity unchanged. The probability simplex has diameter $\sqrt2$, so each centered loss vector has norm at most $G_\ell\sqrt{2N}$. Dudley's entropy integral with the sequence-level normalization $1/N$ gives
\begin{align*}
\widehat{\mathfrak R}_{S_0}(\mathcal Q_k)
&\lesssim G_\ell\sqrt{\frac{\Theta}{N}}
  \int_0^{\sqrt2}
  \sqrt{\log\left(1+\frac{2B_*\Lambda\sqrt{m/k}}{u}\right)}\,du\\
&\lesssim G_\ell\sqrt{\frac{\Theta}{N}}
  \sqrt{1+\log\left(1+2B_*\Lambda\sqrt{m/k}\right)}.
\end{align*}
The last step uses $\int_0^{\sqrt2}\sqrt{\log(1+a/u)}\,du\lesssim\sqrt{1+\log(1+a)}$ for $a\ge0$. Since $k\ge1$, the logarithmic factor is at most a constant times
$1+\log(L+3)+(L+1)\log(2+Hmd)$.
On the other hand, $B_l,G_\pi\ge1$ and $C_l\ge0$ give $\rho_L\ge L$ and $s_L\ge[2(1+Hmd)]^L$, hence
\begin{equation*}
\tau_{m,N}\ge\log\left(1+L\,[2(1+Hmd)]^L\right),
\end{equation*}
which dominates this factor. We have therefore obtained
\begin{equation*}
\widehat{\mathfrak R}_{S_0}(\mathcal Q_k)
\lesssim G_\ell\sqrt{\frac{\Theta\tau_{m,N}}{N}}.
\end{equation*}
Because $dH\ge1$, this also proves the coarser form used in the main text,
\begin{equation*}
\widehat{\mathfrak R}_{S_0}(\mathcal Q_k)
\lesssim G_\ell\sqrt{\frac{\Theta dH\tau_{m,N}}{N}}
+G_\ell\sqrt{\frac{dn_v}{N}}.
\end{equation*}
Here $\Theta$ already counts the head and any trainable embeddings, so the first term alone is valid and the second is redundant; the displayed bound is a looser consequence of the preceding estimate. The proof uses one Rademacher sign per sequence and introduces one factor $G_\ell$; it never replaces these signs by independent signs for individual tokens.

Finally, $N=nT$ gives
\begin{equation*}
T\widehat{\mathfrak R}_{S_0}(\mathcal Q_k)
\lesssim
G_\ell\sqrt{\frac{\Theta dH\tau_{m,nT}T}{n}}
+G_\ell\sqrt{\frac{Tdn_v}{n}}.
\end{equation*}
Substituting this in Eq.~\eqref{eq:erm_final_risk} gives Eq.~\eqref{eq:final_parameterized_ntp}. The coefficients $4T$ and $6B_\ell$ in the ERM bound are unchanged, and the reduction factor $B_\ell k/\log 2$ is applied once. Context length enters through $\tau_{m,nT}$ and also through $\Theta$ when the number of trainable positional-embedding parameters grows with $m$.
\end{proof}

\subsection{Sampling Units}
\label{app:data_dep}

For each step $i$, the dataset $S_i$ contains $n$ independently sampled complete sequences from the corresponding step-level skill distribution $\mathcal D_i$. These sequences need not be extracted from a shared trajectory; each is drawn independently from some real-world instance of the same skill. Dependence among tokens and prefixes within a complete sequence is unrestricted. This independent-record view is the idealization underlying the bounds. The Atom-Task experiments use the one-step and short-chain datasets specified in Section~\ref{sec:experiments}, including three- and five-step chains; the $L$-step accounting serves only as a budget convention, not an independence assumption for transitions within a chain.

\section{Technical appendices and supplementary material}

\begin{table}[ht]
\centering
\caption{Summary of Main Notations}
\label{tab:notation}
\small
\begin{tabularx}{\textwidth}{lX}
\toprule
Symbol & Meaning \\
\midrule
$\mathcal{V}$ & Vocabulary of size $n_v$. \\
$\mathrm{hard}(\cdot)$ & Operator converting probability distributions to discrete tokens via argmax. \\
$\mathcal{D}, \mathcal{D}_i$ & Target problem distribution, and data distribution for the $i$-th reasoning step. \\
$\Psi, \psi, \psi_*$ & Hypothesis class for single-step functions, learned single-step function, and ground-truth mapping (not necessarily within $\Psi$). \\
$\bar{\psi}$ & State transition mapping, $\bar{\psi}(z) := z[k+1:m] \oplus \mathrm{hard}(\psi(z))$, representing the input state for the next step. \\
$\mathcal{G}, \mathcal{H}, g, h$ & Hypothesis classes and functions for the token predictor and representation learner. \\
$\psi_{g,h}$, $F_{\psi_{g,h}}$ & Block predictor of $k$ successive next-token predictions by $g\circ h$, and the resulting final $T$-step predictor. \\
$\ell, \mathfrak{L}$ & Token-level loss ($\epsilon_\ell$-clipped cross-entropy of Assumption~\ref{assum:lipschitz}) and sequence-level loss (average over the $k$ generated tokens). \\
$\epsilon_\ell, G_\ell, B_\ell$ & Clipping threshold of $\ell$ ($\epsilon_\ell\in(0,1/4)$), Lipschitz constant of $\ell$ w.r.t.\ the 2-norm, and bound of $\ell$ (hence of the sequence-level loss $\mathfrak{L}$). \\
$\mathcal{F}$ & Sequence-loss class $\mathcal{F}=\mathfrak{L}\circ\Psi=\{f_\psi:\psi\in\Psi\}$ with $f_\psi(z,\Delta z)=\mathfrak{L}(\psi(z),\Delta z)$. \\
$T, m, k, n$ & Total number of reasoning steps, state length in Section~\ref{pre} (complete-record length in Section~\ref{sec:ntp}), number of generated tokens per step, number of samples per step. \\
$R_{\mathcal{D}}(F_\psi)$ & Final-task population risk, $R_{\mathcal{D}}(F_\psi)=\int \mathfrak{L}(F_\psi(z_0),\Delta z_T)\,dP(z_0,\Delta z_T)$. \\
$\mathcal{D}', P$ & $\mathcal{D}$ is the law of input--answer pairs $(z_0,\Delta z_T)$ with measure $P$, and $\mathcal{D}'$ the joint law of the reasoning trajectory $(z_0,\Delta z_1,\dots,\Delta z_T)$. \\
$R_{\mathcal{D}_i}^k, \widehat{R}_{S_i}^k$ & Target-token population and empirical risks of the block predictor at the $i$-th reasoning step. \\
$N, q_{\psi_{g,h}}, \mathcal Q_k$ & Number of complete sequence samples, target-token loss, and its function class. \\
$Q_i, \operatorname{disc}_{\mathcal F}$ & Atomic training distribution and train--target discrepancy. \\
$\mathcal{E}_k(\widehat\psi)$ & Target-token atomic excess risk of the learned block predictor, where $\widehat\psi:=\psi_{\widehat g,\widehat h}$. \\
$\Theta, L, d, H, n_v$ & Number of model parameters, number of layers, model dimension, number of attention heads, vocabulary size. \\
$\Pi_k, \tau_{m,N}$ & Final-$k$ row selector and decoder-capacity factor. \\
\bottomrule
\end{tabularx}
\end{table}

\subsection{Technical Preliminaries}
\label{app:preliminaries}

We provide the formal definitions for the theoretical frameworks used in our analysis.

\paragraph{PAC Learning}
We conduct analysis within the Probably Approximately Correct (PAC) framework, deriving explicit upper bounds for the generalization error that hold with probability at least $1-\delta$.

\paragraph{Rademacher Complexity}
Rademacher complexity measures the richness of a function class $\mathcal{F}$. For a fixed sample $S=\{x_1, \dots, x_N\}$, the empirical Rademacher complexity and expected Rademacher complexity are defined as
\begin{equation*}
    \widehat{\mathfrak{R}}_S(\mathcal{F}) = \mathbb{E}_{\boldsymbol{\sigma}} \left[ \sup_{f \in \mathcal{F}} \frac{1}{N} \sum_{i=1}^{N} \sigma_i f(x_i) \right], \quad \mathfrak{R}_N(\mathcal{F}) = \mathbb{E}_S[\widehat{\mathfrak{R}}_S(\mathcal{F})],
\end{equation*}
where $\sigma_i$ are independent Rademacher random variables taking values $\pm 1$ with equal probability.

\paragraph{Total Variation Distance}
For two probability distributions $P$ and $Q$ over the same measurable space $\mathcal{X}$, the total variation distance is defined as
\begin{equation*}
    \|P - Q\|_{\mathrm{TV}} = \sup_{A \subseteq \mathcal{X}} |P(A) - Q(A)|.
\end{equation*}
Since $|\mathfrak{L}|\le B_\ell$, this distance bounds the discrepancy of Corollary~\ref{cor:atomic_shift}: $\operatorname{disc}_{\mathcal F}(\mathcal D_i,Q_i)\le 2B_\ell\|\mathcal D_i-Q_i\|_{\mathrm{TV}}$.

\subsection{Local Error Floor of a Hard-Decoding Error}
\label{app:proof_bound_M}

To derive the generalization error decomposition, we first record the loss incurred by a hard-decoding error at a single step.

\begin{lemma}[Local Error Floor of a Hard-Decoding Error]
\label{lem:bound_M}
For the autoregressive predictor $\psi$, every state $s$ and every ground-truth block $\Delta s$ satisfy
\begin{equation*}
    \mathfrak{L}(\psi(s), \Delta s) \ \ge\ \frac{\log 2}{k}\,\ind\{\mathrm{hard}(\psi(s))\neq\Delta s\},
\end{equation*}
where $(\log 2)/k$ is a lower bound on the sequence-level loss $\mathfrak{L}$ whenever the hard decoding differs from the ground truth.
\end{lemma}

\begin{proof}
Fix a state $s$ and a ground-truth block $\Delta s$, and consider a token whose ground-truth index is $p$ and whose argmax index is $q\neq p$. Then $\hat{y}_{q} \ge \hat{y}_{p}$, and since probabilities are non-negative and sum to at most $1$:
\begin{equation*}
    1 \ge \hat{y}_{p} + \hat{y}_{q} \ge 2\hat{y}_{p} \implies \hat{y}_{p} \le 1/2.
\end{equation*}
Since $\max\{\hat{y}_{p},\epsilon_\ell\}\le 1/2$ whenever $\hat{y}_{p}\le 1/2$ and $\epsilon_\ell<1/4$, the token-level clipped cross-entropy of every mispredicted token satisfies $\ell=-\log\max\{\hat{y}_{p},\epsilon_\ell\}\ge\log 2$. A block is hard-decoding-erroneous only if at least one of its $k$ tokens is mispredicted, so averaging over the $k$ tokens yields $\mathfrak{L}(\psi(s),\Delta s)\ge\frac{\log 2}{k}>0$, which is the claim.
\end{proof}
\begin{remark}
The bound uses deterministic hard predictions: if the hard predictions at a step match the ground truth then no error is recorded, and otherwise the deviation is charged through the lower bound $(\log 2)/k$ on the local loss. This is the fact used by the risk decomposition of Theorem~\ref{thm:decomp}.
\end{remark}
\subsection{Proof of Theorem~\ref{thm:decomp} (Generalization Error Decomposition)}
\label{app:proof_decomp}
\begin{proof}
We aim to bound the generalization error $R_{\mathcal{D}}(F_\psi)$. Let $\mathbf z=(z_0,\Delta z_1,\dots,\Delta z_T)\sim\mathcal{D}'$ and let $p(\mathbf z)$ denote its density. Recall that the oracle state is written $z_i$, whereas the free-running (FR) state generated by the model is written with a hat, $\widehat z_0=z_0$ and $\widehat z_i=\bar\psi(\widehat z_{i-1})$, so that
\begin{equation*}
    R_{\mathcal{D}}(F_\psi)=\int \mathfrak{L}\big(\psi(\widehat z_{T-1}),\Delta z_T\big)\,dp(\mathbf z).
\end{equation*}
We first record a fact that will be used repeatedly: \emph{the FR and oracle trajectories coincide exactly up to the first hard-decoding error.} Formally, let
\begin{equation*}
    u_i:=\ind\big\{\mathrm{hard}(\psi(\widehat z_{i-1}))\neq\Delta z_i\big\},
    \qquad
    I:=\inf\{i\ge1:u_i=1\},
\end{equation*}
with $I=\infty$ if no error occurs. Then on the event $\{u_1=\cdots=u_{i-1}=0\}$ the two states agree, $\widehat z_{i-1}=z_{i-1}$. Indeed, $\widehat z_0=z_0$, and if $\widehat z_{j-1}=z_{j-1}$ while $u_j=0$, then
\begin{equation*}
    \widehat{\Delta z}_j=\mathrm{hard}(\psi(\widehat z_{j-1}))
    =\mathrm{hard}(\psi(z_{j-1}))=\Delta z_j,
\end{equation*}
so that
\begin{equation*}
    \widehat z_j
    =\widehat z_{j-1}[k{+}1{:}m]\oplus\widehat{\Delta z}_j
    =z_{j-1}[k{+}1{:}m]\oplus\Delta z_j
    =z_j.
\end{equation*}
The claim follows by induction. We stress that this is a consequence of hard decoding alone---before the first error the two trajectories are \emph{identical} rather than merely close---and that no Lipschitz or stability property of $\psi$ is used.

Next, we split the final-step risk at the first error. On $\{I>T-1\}$ the observation above gives $\widehat z_{T-1}=z_{T-1}$, so the integrand is exactly the step-$T$ oracle loss; on $\{I\le T-1\}$ it is bounded by $B_\ell$. Pointwise,
\begin{equation*}
    \mathfrak{L}\big(\psi(\widehat z_{T-1}),\Delta z_T\big)
    \le \mathfrak{L}\big(\psi(z_{T-1}),\Delta z_T\big)+B_\ell\,\ind\{I\le T-1\},
\end{equation*}
and integrating against $p(\mathbf z)$ yields
\begin{equation*}
    R_{\mathcal{D}}(F_\psi)
    \le R_{\mathcal{D}_T}(\psi)+B_\ell\,\Pr(I\le T-1)
    =R_{\mathcal{D}_T}(\psi)+B_\ell\sum_{i=1}^{T-1}\Pr(I=i),
\end{equation*}
where the last step uses that the events $\{I=i\}$ are disjoint.

It remains to charge each first-error probability to the corresponding atomic risk. Fix $i\le T-1$. Since $\{I=i\}=\{I\ge i\}\cap\{u_i=1\}$, and since on $\{I\ge i\}$ the observation above gives $\widehat z_{i-1}=z_{i-1}$, the event $\{I=i\}$ coincides with
\begin{equation*}
    \{I\ge i\}\cap\big\{\mathrm{hard}(\psi(z_{i-1}))\neq\Delta z_i\big\}.
\end{equation*}
Dropping the event $\{I\ge i\}$, which can only increase the probability, and using that $(z_{i-1},\Delta z_i)$ has law $\mathcal D_i$, we obtain
\begin{equation*}
    \Pr(I=i)\le\Pr_{\mathcal D_i}\big(\mathrm{hard}(\psi(z_{i-1}))\neq\Delta z_i\big).
\end{equation*}
Finally, applying the error floor of Lemma~\ref{lem:bound_M} pointwise and taking the expectation over $\mathcal D_i$ gives
\begin{equation*}
    \Pr_{\mathcal D_i}\big(\mathrm{hard}(\psi(z_{i-1}))\neq\Delta z_i\big)
    \le\frac{k}{\log 2}\,\mathbb E_{\mathcal D_i}\big[\mathfrak{L}(\psi(z_{i-1}),\Delta z_i)\big]
    =\frac{k\,R_{\mathcal{D}_i}(\psi)}{\log 2}.
\end{equation*}
Summing over $i=1,\dots,T-1$ thus gives
\begin{equation*}
    \Pr(I\le T-1)\le\frac{k}{\log 2}\sum_{i=1}^{T-1}R_{\mathcal{D}_i}(\psi).
\end{equation*}

Finally, we combine the two estimates. Since $\ell(p,y)\le\log(1/\epsilon_\ell)$ for all $p,y$, we may take $B_\ell=\log(1/\epsilon_\ell)$; then $\epsilon_\ell<1/4$ and $k\ge1$ imply
\begin{equation*}
    \frac{B_\ell k}{\log 2}\ge 2>1,
\end{equation*}
the single term $R_{\mathcal{D}_T}(\psi)$ is dominated by $\frac{B_\ell k}{\log 2}R_{\mathcal{D}_T}(\psi)$, and hence
\begin{equation*}
    R_{\mathcal{D}}(F_\psi)\le\frac{B_\ell k}{\log 2}\sum_{i=1}^{T}R_{\mathcal{D}_i}(\psi).
\end{equation*}
Because each intermediate state $z_{i-1}$ is deterministically constructed from the prefix $(z_0,\Delta z_1,\dots,\Delta z_{i-1})$, the joint law of $(z_{i-1},\Delta z_i)$ induced by $\mathcal{D}'$ is exactly the sub-task distribution $\mathcal{D}_i$; marginalizing out the remaining variables, the bound becomes
\begin{equation*}
    R_{\mathcal{D}}(F_\psi)\le\frac{B_\ell k}{\log 2}\sum_{i=1}^{T}\int \mathfrak{L}\big(\psi(z_{i-1}),\Delta z_i\big)\,d\mathcal{D}_i(z_{i-1},\Delta z_i),
\end{equation*}
which is the desired decomposition. The coefficient $\frac{B_\ell k}{\log 2}$ is thus depth-independent, with $(\log 2)/k$ the local error floor of Lemma~\ref{lem:bound_M}.
\end{proof}

\subsection{Proof of Theorem~\ref{lem:rad} (Rademacher Complexity Bound)}
\label{app:proof_rad}
We first present the multi-distribution symmetrization lemma used in Section~\ref{sec:rademacher}.
\begin{lemma}
\label{lem:symm}
Define the population risk $R_{\mathcal{D}_i}(\psi):=\int \mathfrak{L}(\psi(z_{i-1}),\Delta z_i)d\mathcal{D}_i(z_{i-1},\Delta z_i)$ and empirical risk $\widehat{R}_{S_i}(\psi):=\frac{1}{n}\sum_{(z_{i-1},\Delta z_i)\in S_i}\mathfrak{L}(\psi(z_{i-1}), \Delta z_i)$. We have:
\begin{equation*}
\mathbb{E}_{\{S_j \sim \mathcal{D}_j^n\}_{j=1}^T} \left[ \sup\nolimits_{\psi\in \Psi} \left( \sum\nolimits_{i=1}^{T} (R_{\mathcal{D}_i}(\psi)-\widehat{R}_{S_i}(\psi)) \right) \right] \leq 2T \ \mathbb{E}_{\{S_j \sim \mathcal{D}_j^n\}_{j=1}^T}[\widehat{\mathfrak{R}}_{S_0}(\mathcal{F})],
\end{equation*}
where $\mathcal{F} = \mathfrak{L} \circ \Psi$, $S_0 = \bigcup_{i=1}^T S_i$, and $\widehat{\mathfrak{R}}_{S_0}(\mathcal{F})$ is the empirical Rademacher complexity of $\mathcal{F}$ on $S_0$.
\end{lemma}
We now prove Lemma~\ref{lem:symm}.
\begin{proof}
Let $S_1', S_2', \dots, S_T'$ be another set of i.i.d. samples, where $S_i' \sim \mathcal{D}_i^n$ and 
\[S_i'=\{(x_{i1}', y_{i1}'), (x_{i2}', y_{i2}'), \dots, (x_{in}', y_{in}')\}\quad\text{for }i=1,2,\dots,T.\]
 Clearly, for all $\psi \in \Psi$, the population risk is the expectation of the empirical risk over the ghost sample:
\begin{equation*}
R_{\mathcal{D}_i}(\psi) = \mathbb{E}_{\{S_j' \sim \mathcal{D}_j^n\}_{j=1}^T}[\hat{R}_{S_i'}(\psi)].
\end{equation*}
Therefore, for every $\psi \in \Psi$, we have
\begin{equation*}
R_{\mathcal{D}_i}(\psi) - \widehat{R}_{S_i}(\psi) = \mathbb{E}_{\{S_j' \sim \mathcal{D}_j^n\}_{j=1}^T}[\hat{R}_{S_i'}(\psi) - \widehat{R}_{S_i}(\psi)].
\end{equation*}
Taking the supremum over $\psi \in \Psi$ on both sides, and using the fact that the supremum of an expectation is upper bounded by the expectation of the supremum (Jensen's inequality), we obtain
\begin{equation*}
\begin{split}
    &\sup_{\psi\in \Psi} \left( \sum_{i=1}^{T} (R_{\mathcal{D}_i}(\psi)-\widehat{R}_{S_i}(\psi)) \right) \\
    =& \sup_{\psi\in \Psi} \mathbb{E}_{\{S_j' \sim \mathcal{D}_j^n\}_{j=1}^T} \left[ \sum_{i=1}^{T} (\hat{R}_{S_i'}(\psi)-\widehat{R}_{S_i}(\psi)) \right] \\
    \le& \mathbb{E}_{\{S_j' \sim \mathcal{D}_j^n\}_{j=1}^T} \left[ \sup_{\psi\in \Psi} \left( \sum_{i=1}^{T} (\hat{R}_{S_i'}(\psi)-\widehat{R}_{S_i}(\psi)) \right) \right].
\end{split}
\end{equation*}
Taking expectation over $\{S_j \sim \mathcal{D}_j^n\}_{j=1}^T$ on both sides yields
\begin{equation*}
\begin{split}
    &\mathbb{E}_{\{S_j \sim \mathcal{D}_j^n\}_{j=1}^T} \left[ \sup_{\psi\in \Psi} \left( \sum_{i=1}^{T} (R_{\mathcal{D}_i}(\psi)-\widehat{R}_{S_i}(\psi)) \right) \right] \\
    \le& \mathbb{E}_{\{S_j \sim \mathcal{D}_j^n\}_{j=1}^T, \{S_j' \sim \mathcal{D}_j^n\}_{j=1}^T} \left[ \sup_{\psi\in \Psi} \left( \sum_{i=1}^{T} (\hat{R}_{S_i'}(\psi)-\widehat{R}_{S_i}(\psi)) \right) \right] \\
    =&\frac{1}{n} \mathbb{E}_{\{S_j \sim \mathcal{D}_j^n\}_{j=1}^T, \{S_j' \sim \mathcal{D}_j^n\}_{j=1}^T} \left[ \sup_{\psi\in \Psi} \left( \sum_{i=1}^{T} \sum_{k=1}^{n} (\mathfrak{L}(\psi(x_{ik}'), y_{ik}')-\mathfrak{L}(\psi(x_{ik}), y_{ik})) \right) \right].
\end{split}
\end{equation*}

Next, we note that for each $i$ and $k$, $(x_{ik}, y_{ik})$ and $(x_{ik}', y_{ik}')$ are i.i.d. variables. Therefore, we can replace them without affecting the expectation. For any specific pair $(i,k)$, let the summation over the remaining indices be denoted by $\sum_{(i',k') \neq (i,k)}$. We have:
\begin{equation*}
\begin{split}
    &\mathbb{E}_{\{S_j \sim \mathcal{D}_j^n\}_{j=1}^T, \{S_j' \sim \mathcal{D}_j^n\}_{j=1}^T} \Bigg[ \sup_{\psi\in \Psi} \Bigg( (\mathfrak{L}(\psi(x_{ik}'), y_{ik}')-\mathfrak{L}(\psi(x_{ik}), y_{ik})) \\
    &\quad + \sum_{(i',k') \neq (i,k)} (\mathfrak{L}(\psi(x_{i'k'}'), y_{i'k'}')-\mathfrak{L}(\psi(x_{i'k'}), y_{i'k'})) \Bigg) \Bigg] \\
    =&\mathbb{E}_{\{S_j \sim \mathcal{D}_j^n\}_{j=1}^T, \{S_j' \sim \mathcal{D}_j^n\}_{j=1}^T} \Bigg[ \sup_{\psi\in \Psi} \Bigg( (-\mathfrak{L}(\psi(x_{ik}'), y_{ik}')+\mathfrak{L}(\psi(x_{ik}), y_{ik})) \\
    &\quad + \sum_{(i',k') \neq (i,k)} (\mathfrak{L}(\psi(x_{i'k'}'), y_{i'k'}')-\mathfrak{L}(\psi(x_{i'k'}), y_{i'k'})) \Bigg) \Bigg].
\end{split}
\end{equation*}

Let $\sigma_{ik}$ be a random variable such that $\mathbb{P}[\sigma_{ik}=1]=\mathbb{P}[\sigma_{ik}=-1]=\frac{1}{2}$. Combining the two equalities above, we obtain:
\begin{equation*}
\begin{split}
    &\mathbb{E}_{\{S_j \sim \mathcal{D}_j^n\}_{j=1}^T, \{S_j' \sim \mathcal{D}_j^n\}_{j=1}^T} \Bigg[ \sup_{\psi\in \Psi} \Bigg( (\mathfrak{L}(\psi(x_{ik}'), y_{ik}')-\mathfrak{L}(\psi(x_{ik}), y_{ik})) \\
    &\quad + \sum_{(i',k') \neq (i,k)} (\mathfrak{L}(\psi(x_{i'k'}'), y_{i'k'}')-\mathfrak{L}(\psi(x_{i'k'}), y_{i'k'})) \Bigg) \Bigg] \\
    =& \mathbb{E}_{\{S_j \sim \mathcal{D}_j^n\}_{j=1}^T, \{S_j' \sim \mathcal{D}_j^n\}_{j=1}^T} \Bigg[ \mathbb{E}_{\sigma_{ik}} \sup_{\psi\in \Psi} \Bigg( \sigma_{ik}(\mathfrak{L}(\psi(x_{ik}'), y_{ik}')-\mathfrak{L}(\psi(x_{ik}), y_{ik})) \\
    &\quad + \sum_{(i',k') \neq (i,k)} (\mathfrak{L}(\psi(x_{i'k'}'), y_{i'k'}')-\mathfrak{L}(\psi(x_{i'k'}), y_{i'k'})) \Bigg) \Bigg].
\end{split}
\end{equation*}
The inner expectation over $\sigma_{ik}$ splits the right-hand side into the two half-terms carrying signs $+$ and $-$; the swap symmetry of the i.i.d.\ pair $(x_{ik},y_{ik})\leftrightarrow(x_{ik}',y_{ik}')$ used in the preceding display shows that these two half-terms are equal, so the identity holds after averaging over $\sigma_{ik}$.
Repeating this process for all $i=1,\dots,T$ and $k=1,\dots,n$, we have
\begin{equation*}
\begin{split}
    &\mathbb{E}_{\{S_j \sim \mathcal{D}_j^n\}_{j=1}^T, \{S_j' \sim \mathcal{D}_j^n\}_{j=1}^T} \left[ \sup_{\psi\in \Psi} \left( \sum_{i=1}^{T} \sum_{k=1}^{n} (\mathfrak{L}(\psi(x_{ik}'), y_{ik}')-\mathfrak{L}(\psi(x_{ik}), y_{ik})) \right) \right] \\
    =& \mathbb{E}_{\substack{\{S_j \sim \mathcal{D}_j^n\}_{j=1}^T,\ \{S_j' \sim \mathcal{D}_j^n\}_{j=1}^T,\\ \{\{\sigma_{ik}\}_{i=1}^{T}\}_{k=1}^{n}}} \left[ \sup_{\psi\in \Psi} \left( \sum_{i=1}^{T} \sum_{k=1}^{n} \sigma_{ik} (\mathfrak{L}(\psi(x_{ik}'), y_{ik}')-\mathfrak{L}(\psi(x_{ik}), y_{ik})) \right) \right].
\end{split}
\end{equation*}

Finally, applying the triangle inequality to the supremum:
\begin{equation*}
\begin{split}
    &\sup_{\psi\in \Psi} \sum_{i=1}^T \sum_{k=1}^n \sigma_{ik}(\mathfrak{L}(\psi(x_{ik}'), y_{ik}')-\mathfrak{L}(\psi(x_{ik}), y_{ik})) \\
    \le& \sup_{\psi\in \Psi} \sum_{i=1}^T \sum_{k=1}^n \sigma_{ik} \mathfrak{L}(\psi(x_{ik}'), y_{ik}') \\
    &+ \sup_{\psi\in \Psi} \sum_{i=1}^T \sum_{k=1}^n -\sigma_{ik} \mathfrak{L}(\psi(x_{ik}), y_{ik}).
\end{split}
\end{equation*}
The joint distribution of the Rademacher signs is invariant under sign reversal. Taking expectations in the preceding inequality and using this symmetry for the second term, we get
\begin{equation*}
\begin{split}
    &\mathbb{E}_{\{S_j \sim \mathcal{D}_j^n\}_{j=1}^T, \{S_j' \sim \mathcal{D}_j^n\}_{j=1}^T, \{\{\sigma_{ik}\}_{i=1}^{T}\}_{k=1}^{n}} \left[ \sup_{\psi\in \Psi} \sum_{i=1}^T \sum_{k=1}^n \sigma_{ik}(\mathfrak{L}(\psi(x_{ik}'), y_{ik}')-\mathfrak{L}(\psi(x_{ik}), y_{ik})) \right] \\
    \le& \mathbb{E}_{\{S_j \sim \mathcal{D}_j^n\}_{j=1}^T, \{S_j' \sim \mathcal{D}_j^n\}_{j=1}^T, \{\{\sigma_{ik}\}_{i=1}^{T}\}_{k=1}^{n}} \left[ \sup_{\psi\in \Psi} \sum_{i=1}^T \sum_{k=1}^n \sigma_{ik} \mathfrak{L}(\psi(x_{ik}'), y_{ik}') \right] \\
    &+ \mathbb{E}_{\{S_j \sim \mathcal{D}_j^n\}_{j=1}^T, \{S_j' \sim \mathcal{D}_j^n\}_{j=1}^T, \{\{\sigma_{ik}\}_{i=1}^{T}\}_{k=1}^{n}} \left[ \sup_{\psi\in \Psi} \sum_{i=1}^T \sum_{k=1}^n \sigma_{ik} \mathfrak{L}(\psi(x_{ik}), y_{ik}) \right] \\
    =& 2 \cdot \mathbb{E}_{\{S_j \sim \mathcal{D}_j^n\}_{j=1}^T, \{S_j' \sim \mathcal{D}_j^n\}_{j=1}^T} \left[ nT \cdot \widehat{\mathfrak{R}}_{S_0}(\mathcal{F}) \right] = 2nT \cdot \mathbb{E}_{\{S_j \sim \mathcal{D}_j^n\}_{j=1}^T}[\widehat{\mathfrak{R}}_{S_0}(\mathcal{F})].
\end{split}
\end{equation*}
Substituting this result into the previous equation involving the $1/n$ factor, we obtain the final bound:
$$
\begin{aligned}
&\mathbb{E}_{\{S_j \sim \mathcal{D}_j^n\}_{j=1}^T}
\left[\sup_{\psi\in\Psi}\left(\sum_{i=1}^{T}
(R_{\mathcal D_i}(\psi)-\hat R_{S_i}(\psi))\right)\right]\\
&\quad\leq \frac{1}{n}\cdot2nT\mathbb{E}_{\{S_j \sim \mathcal D_j^n\}_{j=1}^T}
[\widehat{\mathfrak{R}}_{S_0}(\mathcal{F})]\\
&\quad=2T\mathbb{E}_{\{S_j \sim \mathcal D_j^n\}_{j=1}^T}
[\widehat{\mathfrak{R}}_{S_0}(\mathcal{F})].
\end{aligned}
$$
\end{proof}

\begin{lemma}[McDiarmid's Inequality\cite{shalev2014understanding}]
\label{lem:mcdiarmid}
Let $V$ be a domain, and let $f: V^{n} \to \mathbb{R}$ be a function of $n$ variables satisfying for some $c>0$, for all $i \in [n]$ and all $x_1, \dots, x_n, x_i' \in V$:
$$
\left| f(x_1, \ldots, x_i, \ldots, x_n) - f(x_1, \ldots, x_i', \ldots, x_n) \right| \leq c.
$$
Let $X_1, X_2, \dots, X_n$ be independent random variables taking values in $V$. Then with probability at least $1-\delta$:
$$
\Big| f(X_1, \ldots, X_n) - \mathbb{E}\big[ f(X_1, \ldots, X_n) \big] \Big| \leq c \sqrt{\frac{n}{2} \ln \left( \frac{2}{\delta} \right)}.
$$
\end{lemma}

Now we give the proof of Theorem~\ref{lem:rad}.

\textbf{Theorem~\ref{lem:rad} (Rademacher Complexity Bound).}
1. Under the preceding conditions (where the step-level loss is bounded by $B_\ell$), with probability at least $1-\delta$:
$$
\sum_{i=1}^{T} (R_{\mathcal{D}_i}(\psi)-\widehat{R}_{S_i}(\psi)) \leq 2T\mathbb{E}_{\{S_j \sim \mathcal{D}_j^n\}_{j=1}^T}[\widehat{\mathfrak{R}}_{S_0}(\mathcal{F})] + B_\ell\sqrt{\frac{2T\log\frac{2}{\delta}}{n}}.
$$
2. Under the preceding conditions (where the step-level loss is bounded by $B_\ell$), with probability at least $1-\delta$:
$$
\left|\sum_{i=1}^{T} (R_{\mathcal{D}_i}(\psi)-\widehat{R}_{S_i}(\psi))\right| \leq 2T\widehat{\mathfrak{R}}_{S_0}(\mathcal{F}) + 3B_\ell\sqrt{\frac{2T\log\frac{4}{\delta}}{n}}.
$$

\begin{proof}
We first prove the first inequality. Since $\sum_{i=1}^{T} (R_{\mathcal{D}_i}(\psi)-\widehat{R}_{S_i}(\psi))$ satisfies the conditions of Lemma~\ref{lem:mcdiarmid} with $Tn$ variables and constant $2B_\ell/n$, applying Lemma~\ref{lem:mcdiarmid} and then Lemma~\ref{lem:symm} yields that with probability at least $1-\delta$:
\begin{equation*}
    \begin{split}
 &\sum_{i=1}^{T} (R_{\mathcal{D}_i}(\psi)-\widehat{R}_{S_i}(\psi)) \\
\le& \sup_{\psi\in\Psi}\left(\sum_{i=1}^{T} (R_{\mathcal{D}_i}(\psi)-\widehat{R}_{S_i}(\psi))\right)\\
\le& \mathbb{E}_{\{S_j \sim \mathcal{D}_j^n\}_{j=1}^T} \left[ \sup_{\psi\in \Psi} \left( \sum_{i=1}^{T} (R_{\mathcal{D}_i}(\psi)-\widehat{R}_{S_i}(\psi)) \right) \right] + B_\ell \sqrt{\frac{2T\log\frac{2}{\delta}}{n}} \\
\le& 2T \mathbb{E}_{\{S_j \sim \mathcal{D}_j^n\}_{j=1}^T}[\widehat{\mathfrak{R}}_{S_0}(\mathcal{F})] + B_\ell \sqrt{\frac{2T\log\frac{2}{\delta}}{n}}.
    \end{split}
\end{equation*}

For the second inequality, based on the first one, since $\widehat{\mathfrak{R}}_{S_0}(\mathcal{F})$ also satisfies the conditions of Lemma~\ref{lem:mcdiarmid} (with $Tn$ variables, constant $2B_\ell/(Tn)$), with probability at least $1-2\delta$ we have:
\begin{equation*}
    \begin{split}
 &\sum_{i=1}^{T} (R_{\mathcal{D}_i}(\psi)-\widehat{R}_{S_i}(\psi)) \\
\le& 2T \mathbb{E}_{\{S_j \sim \mathcal{D}_j^n\}_{j=1}^T}[\widehat{\mathfrak{R}}_{S_0}(\mathcal{F})] + B_\ell \sqrt{\frac{2T\log\frac{2}{\delta}}{n}} \\
\le& 2T \widehat{\mathfrak{R}}_{S_0}(\mathcal{F}) + 2T \left( \frac{2B_\ell}{nT} \sqrt{ \frac{nT\log\frac{2}{\delta}}{2}} \right) + B_\ell \sqrt{\frac{2T\log\frac{2}{\delta}}{n}} \\
=& 2T \widehat{\mathfrak{R}}_{S_0}(\mathcal{F}) + 3B_\ell \sqrt{\frac{2T\log\frac{2}{\delta}}{n}}.
    \end{split}
\end{equation*}
The reverse deviation follows by the same sign-symmetry argument. Using $\delta/2$ in place of $\delta$ converts the $1-2\delta$ event and the $\log(2/\delta)$ factors above into the $1-\delta$ event and the $\log(4/\delta)$ factors of part~2, without changing the coefficients.
\end{proof}

\subsection{Proof of Corollary~\ref{cor:ood_bound}}
\label{app:proof_ood_bound}
\begin{proof}
Let $U:=2T\widehat{\mathfrak R}_{S_0}(\mathcal F)+3B_\ell\sqrt{2T\log(4/\delta)/n}$, so that Theorem~\ref{lem:rad} bounds $\big|\sum_{i}(R_{\mathcal D_i}(\psi)-\widehat R_{S_i}(\psi))\big|$ by $U$ uniformly over $\psi$, on an event of probability at least $1-\delta$.

\emph{First display.} Theorem~\ref{thm:decomp} gives $R_{\mathcal D}(F_\psi)\le\frac{B_\ell k}{\log 2}\sum_{i}R_{\mathcal D_i}(\psi)$; replacing the population risks by their empirical counterparts costs at most $U$, and multiplying by $B_\ell k/\log 2$ gives the first display.

\emph{Second display.} Let $\widehat\psi$ be the sub-task ERM and $\psi^\star$ a population minimizer. Empirical optimality gives $\sum_{i}\widehat R_{S_i}(\widehat\psi)\le\sum_{i}\widehat R_{S_i}(\psi^\star)$, so the atomic excess of $\widehat\psi$ obeys
\begin{align*}
\sum_{i}R_{\mathcal D_i}(\widehat\psi)-\sum_{i}R_{\mathcal D_i}(\psi^\star)
&=\underbrace{\sum_{i}\big(R_{\mathcal D_i}(\widehat\psi)-\widehat R_{S_i}(\widehat\psi)\big)}_{\le U}
+\underbrace{\sum_{i}\big(\widehat R_{S_i}(\widehat\psi)-\widehat R_{S_i}(\psi^\star)\big)}_{\le 0}\\
&\quad+\underbrace{\sum_{i}\big(\widehat R_{S_i}(\psi^\star)-R_{\mathcal D_i}(\psi^\star)\big)}_{\le U}
\le 2U .
\end{align*}
Substituting $\sum_{i}R_{\mathcal D_i}(\widehat\psi)\le\sum_{i}R_{\mathcal D_i}(\psi^\star)+2U$ into the bound of Theorem~\ref{thm:decomp} gives the second display.
\end{proof}

\subsection{Proof of Corollary~\ref{cor:atomic_shift}}
\label{app:proof_shift}
\begin{proof}
By the oracle-prefix reduction of Theorem~\ref{thm:decomp} the final-task risk is bounded by $\frac{B_\ell k}{\log 2}\sum_{i=1}^{T}R_{\mathcal D_i}(\psi)$; transferring each term via $R_{\mathcal D_i}(\psi)\le R_{Q_i}(\psi)+\operatorname{disc}_{\mathcal F}(\mathcal D_i,Q_i)$ and applying Theorem~\ref{lem:rad} to the $Q_i$ proves the first display. The second follows from the same comparison with the deviation paid once for each of $\widehat\psi$ and $\psi^\star$, and the transfer from $Q_i$ to $\mathcal D_i$ likewise paid twice.
\end{proof}

\subsection{Teacher-Forced NTP Reduction}
\label{app:proof_ntp_tf_reduction}

\begin{lemma}[Teacher-Forced NTP Reduction]
\label{lem:ntp_tf_reduction}
Under Assumption~\ref{assum:lipschitz}, consider the NTP setting of Section~\ref{sec:ntp_form}, with deterministic argmax decoding and the same context-window rule for teacher-forced and free-running prefixes. Then
\begin{equation*}
    R_{\mathcal D}(F_{\psi_{g,h}})\le\frac{B_\ell k}{\log 2}\sum_{i=1}^{T}R_{\mathcal D_i}^{k}(\psi_{g,h}).
\end{equation*}
The risks on the right are the teacher-forced target-token risks defined in Eq.~\eqref{eq:target_token_loss}, whereas the risk on the left is evaluated on the free-running final prediction.
\end{lemma}

\begin{proof}
We follow the first-error argument of Theorem~\ref{thm:decomp}, now locating the first erroneous token within the first erroneous block. Fix $g,h$, and let $\psi:=\psi_{g,h}$ be the block predictor given by $k$ successive free-running next-token predictions via $g\circ h$. As in Appendix~\ref{app:proof_decomp}, let $\mathbf z=(z_0,\Delta z_1,\dots,\Delta z_T)\sim\mathcal D'$ with joint law $p(\mathbf z)$, and write
\begin{equation*}
    \widehat z_0=z_0,\qquad
    \widehat{\Delta z}_i=\mathrm{hard}(\psi(\widehat z_{i-1})),\qquad
    \widehat z_i=\bar\psi(\widehat z_{i-1}).
\end{equation*}
For $r=1,\dots,k$, let $X_{i,<r}$ be the oracle context before token $(\Delta z_i)_r$, formed from $z_{i-1}$ and the true within-step prefix $(\Delta z_i)_{<r}$. Let $\widehat X_{i,<r}$ be its free-running counterpart, formed from $\widehat z_{i-1}$ and $(\widehat{\Delta z}_i)_{<r}$. Both contexts use the same context-window rule, including truncation. For each oracle block, abbreviate its teacher-forced target-token loss by
\begin{equation*}
    q_i:=\frac{1}{k}\sum_{r=1}^{k}
    \ell\!\left(g(h(X_{i,<r})),(\Delta z_i)_r\right).
\end{equation*}
This is Eq.~\eqref{eq:target_token_loss} evaluated on the oracle record for step $i$; consequently, $\int q_i\,dp(\mathbf z)=R_{\mathcal D_i}^{k}(\psi_{g,h})$.

Define, as in the proof of Theorem~\ref{thm:decomp},
\begin{equation*}
    u_i:=\ind\big\{\widehat{\Delta z}_i\neq\Delta z_i\big\},\qquad
    I:=\inf\{i\in\{1,\dots,T\}:u_i=1\},
\end{equation*}
with $I=\infty$ if no error occurs. The same induction gives $\widehat z_{i-1}=z_{i-1}$ on $\{I\ge i\}$. On $\{I=i\}$, let $r$ be the first position in block $i$ with $(\widehat{\Delta z}_i)_r\neq(\Delta z_i)_r$. Then
\begin{equation*}
    \widehat z_{i-1}=z_{i-1},\qquad
    (\widehat{\Delta z}_i)_{<r}=(\Delta z_i)_{<r},\qquad
    \widehat X_{i,<r}=X_{i,<r}.
\end{equation*}
Thus the erroneous hard prediction at this position is the argmax of the same probability vector $g(h(X_{i,<r}))$ that appears in the teacher-forced loss. The probability assigned to the true token is therefore at most $1/2$. By the clipped cross-entropy calculation in Lemma~\ref{lem:bound_M},
\begin{equation*}
    \ell\!\left(g(h(X_{i,<r})),(\Delta z_i)_r\right)\ge\log 2,
    \qquad q_i\ge\frac{\log 2}{k}
    \quad\text{on }\{I=i\}.
\end{equation*}
This argument includes $i=T$, so it also accounts for errors within the final output block.

Next, split the final-step loss according to whether any hard error occurs. On $\{I>T\}$, all generated tokens agree with their oracle counterparts, including every prefix inside the final block, and hence
\begin{equation*}
    \mathfrak L\big(\psi(\widehat z_{T-1}),\Delta z_T\big)=q_T.
\end{equation*}
On $\{I=i\}$, boundedness gives $\mathfrak L(\psi(\widehat z_{T-1}),\Delta z_T)\le B_\ell$. Since $B_\ell k/\log 2\ge1$ and the events $\{I>T\},\{I=1\},\dots,\{I=T\}$ form a partition, pointwise we obtain
\begin{align*}
    \mathfrak L\big(\psi(\widehat z_{T-1}),\Delta z_T\big)
    &\le \ind\{I>T\}\,q_T+B_\ell\sum_{i=1}^{T}\ind\{I=i\}\\
    &\le\frac{B_\ell k}{\log 2}\left[
        \ind\{I>T\}\,q_T+\sum_{i=1}^{T}\ind\{I=i\}\,q_i
    \right]\\
    &\le\frac{B_\ell k}{\log 2}\sum_{i=1}^{T}q_i.
\end{align*}
Finally, integrating against $p(\mathbf z)$ and using the oracle step marginals $\mathcal D_i$ yields
\begin{equation*}
    R_{\mathcal D}(F_{\psi_{g,h}})
    \le\frac{B_\ell k}{\log 2}\sum_{i=1}^{T}\int q_i\,dp(\mathbf z)
    =\frac{B_\ell k}{\log 2}\sum_{i=1}^{T}R_{\mathcal D_i}^{k}(\psi_{g,h}),
\end{equation*}
which proves the claim. The argument uses equality of the two prefixes only before the first hard error; it does not identify the teacher-forced and free-running block losses after that error.
\end{proof}

\subsection{Proof of Theorem~\ref{thm:ntp}}
\label{app:proof_ntp}
\begin{proof}
Apply Theorem~\ref{lem:rad} to the bounded sequence-level class $\mathcal Q_k$ and the samples $S_i\sim\mathcal D_i^n$. Uniformly over $g\in\mathcal G$ and $h\in\mathcal H$,
\begin{equation*}
\left|\sum_{i=1}^{T}\left(R_{\mathcal D_i}^{k}(\psi_{g,h})-\widehat R_{S_i}^{k}(\psi_{g,h})\right)\right|
\le 2T\widehat{\mathfrak R}_{S_0}(\mathcal Q_k)+3B_\ell\sqrt{\frac{2T\log(4/\delta)}{n}},
\end{equation*}
which is Eq.~\eqref{eq:target_ntp_bound}.

For the ERM comparison, the three-term decomposition of the proof of Corollary~\ref{cor:ood_bound} (Appendix~\ref{app:proof_ood_bound}) applies verbatim with $\mathcal F$ replaced by $\mathcal Q_k$ and $R_{\mathcal D_i}$ by $R_{\mathcal D_i}^{k}$: writing $\mathcal E_k(\widehat\psi):=\sum_{i}R_{\mathcal D_i}^{k}(\widehat\psi)-\sum_{i}R_{\mathcal D_i}^{k}(\psi_{g^\star,h^\star})$ for the target-token atomic excess and $U_k:=2T\widehat{\mathfrak R}_{S_0}(\mathcal Q_k)+3B_\ell\sqrt{2T\log(4/\delta)/n}$, empirical optimality and the two directions of the same uniform event give $\mathcal E_k(\widehat\psi)\le U_k+0+U_k$. Lemma~\ref{lem:ntp_tf_reduction} then gives
\begin{equation*}
R_{\mathcal D}(F_{\psi_{\widehat g,\widehat h}})\le\frac{B_\ell k}{\log 2}\sum_{i=1}^{T}R_{\mathcal D_i}^{k}(\widehat\psi)=\frac{B_\ell k}{\log 2}\left[\sum_{i=1}^{T}R_{\mathcal D_i}^{k}(\psi_{g^\star,h^\star})+\mathcal E_k(\widehat\psi)\right],
\end{equation*}
which is Eq.~\eqref{eq:erm_final_risk} after substituting $U_k+0+U_k=4T\widehat{\mathfrak R}_{S_0}(\mathcal Q_k)+6B_\ell\sqrt{2T\log(4/\delta)/n}$. The projection and loss-cover argument in Appendix~\ref{app:dim_derivation} gives Eq.~\eqref{eq:qk_capacity}.
\end{proof}

\subsection{Tightness of the Additive Reduction}
\label{app:proof_additive_tightness}
\begin{proposition}[Tightness of the Additive Reduction]
\label{prop:additive_tightness}
For every $T\ge1$ and every $p\in[0,1/T]$ there is a task with the following complete description. Let $\mathcal V=\{0,1\}$, $k=1$, $m=T+1$, and draw $u_1,\dots,u_T$ i.i.d.\ $\operatorname{Bernoulli}(p)$. The oracle trajectory is $z_0=(u_1,\dots,u_T,0)$ with $\Delta z_i=0$ for every $i$, so that the oracle state before step $i$ is $z_{i-1}=(u_i,\dots,u_T,0,\dots,0)$, whose first token is $u_i$ and whose last token is $0$, and the step marginal $\mathcal D_i$ is the law of $(z_{i-1},0)$ induced by $(u_i,\dots,u_T)$. The shared single-step predictor is the deterministic rule $\mathrm{hard}(\psi(z))=\max\{(z)_1,(z)_m\}$, with $\psi(z)$ putting probability $1$ on that token, so that the clipped cross-entropy of Assumption~\ref{assum:lipschitz} charges $0$ when the output is correct and $B_\ell=\log(1/\epsilon_\ell)$ otherwise. Writing $e_0=0$ and $e_i=e_{i-1}\vee u_i$ for the error flag ($\vee$ is logical OR, and a local error puts the state in an absorbing error mode), this task satisfies
\begin{equation*}
R_{\mathcal D}(F_\psi)
=B_\ell\left[1-(1-p)^T\right]
=\Omega\!\left(\sum_{i=1}^{T}R_{\mathcal D_i}(\psi)\right),
\qquad
\sum_{i=1}^{T}R_{\mathcal D_i}(\psi)=TpB_\ell .
\end{equation*}
\end{proposition}

\begin{proof}
Verify the construction of the statement. On an oracle state the last token is $0$, so the hard output equals $u_i$ and the error event is $u_i=1$; hence $R_{\mathcal D_i}(\psi)=pB_\ell$. Once an error occurs, its output $1$ becomes the last token of every later state and the rule keeps emitting $1$, so the final output is wrong exactly when $\bigvee_{i=1}^{T}u_i=1$, which gives
\begin{equation*}
R_{\mathcal D}(F_\psi)
=B_\ell\Pr\!\left(\bigvee_{i=1}^{T}u_i=1\right)
=B_\ell\left[1-(1-p)^T\right].
\end{equation*}
The union bound gives $1-(1-p)^T\le Tp$, and for $Tp\le1$,
$1-(1-p)^T\ge Tp-\binom{T}{2}p^2\ge\frac12Tp$. Consequently
\begin{equation*}
\frac12TpB_\ell
\le R_{\mathcal D}(F_\psi)
\le TpB_\ell
=\sum_{i=1}^{T}R_{\mathcal D_i}(\psi),
\end{equation*}
hence $R_{\mathcal D}(F_\psi)=\Omega\big(\sum_{i=1}^{T}R_{\mathcal D_i}(\psi)\big)$, which proves the order-tightness of the additive reduction.
\end{proof}

\paragraph{What the construction shows.} The witness combines two features: the local error events are independent across steps, and the state carries an absorbing error flag, so that the final output is wrong exactly when at least one local error has occurred. That event has probability $1-(1-p)^T=\Theta(Tp)$ for $Tp\le1$, of the same order as the sum of the $T$ step risks, while the per-step risk itself is $R_{\mathcal D_i}(\psi)=pB_\ell$. The additive form of Theorem~\ref{thm:decomp} is therefore tight up to its depth-independent coefficient, and the linear-in-$T$ growth of the bound---here the genuine linear growth of the target risk---cannot be removed by any refinement of the argument.

\subsection{Tightness of the Atomic Excess}
\label{app:proof_pooled_lower}
\begin{proposition}[Tightness of the Atomic Excess]
\label{prop:pooled_erm_lower}
Fix $T=4r+1$ with $r\in\mathbb N$, $r\ge1$, a sign $\theta\in\{-1,+1\}$, and $\gamma\in(0,1/(2T)]$; the sign $\theta$ only selects which of the two hypotheses is favoured, and the constants below do not depend on it. Let $\mathcal V=\{0,1\}$, $k=1$, $m=T+1$, and let $P_\theta$ be the law on $\{0,1\}^T$ that assigns weight $\frac{1}{2T}+\theta\gamma$ to each vector $u^{(j)}\in\{0,1\}^T$, $j\in[T]$, whose support is the cyclic block $\{j,j+1,\dots,j+2r\}$ of $2r+1$ consecutive coordinates (indices modulo $T$), and weight $\frac{1}{2T}-\theta\gamma$ to $v^{(j)}$, the same block with its leading coordinate $j$ flipped, so that $v^{(j)}$ has ones exactly on $\{j+1,\dots,j+2r\}$; these $2T$ weights sum to one. For $X\sim P_\theta$, $P_\theta(X_i=1)=\frac12+\theta\gamma$ and $P_\theta\big(\bigoplus_{i=1}^{T}X_i=1\big)=\frac12+\theta T\gamma$, where $\bigoplus$ denotes bitwise XOR in this construction (not sequence concatenation). The oracle trajectory is $z_0=(X_1,\dots,X_T,0)$ with $\Delta z_i=p_i$ and $p_i=p_{i-1}\oplus X_i$, $p_0=0$, so that the state before step $i$ exposes $X_i$ as its first token and $p_{i-1}$ as its last token, and $\mathcal D_i^\theta$ is the law of the $i$-th oracle state and update. The two-element class is $\Psi=\{\psi_0,\psi_1\}$, with $\psi_s(z)$ putting probability $3/4$ on token $(z)_m\oplus s$ and $1/4$ on the other token, so that a correct output costs $\ell_{\mathrm{ok}}=\log(4/3)$ and a wrong one $\ell_{\mathrm{err}}=\log4$ (clipping is inactive since $\epsilon_\ell<1/4$). With $n$ i.i.d.\ draws from each $\mathcal D_i^\theta$, the sub-task ERM $\widehat\psi=\arg\min_{\psi\in\Psi}\sum_{i=1}^{T}\widehat R_{S_i}(\psi)$ satisfies, for $\gamma=c_0\min\{1/\sqrt{nT},1/T\}$ with a universal constant $c_0\le1/2$,
\begin{equation*}
\mathbb E_{S}\!\left[\sum_{i=1}^{T}R_{\mathcal D_i}^{\theta}(\widehat\psi)-\min_{\psi\in\Psi}\Bigl\{\sum_{i=1}^{T}R_{\mathcal D_i}^{\theta}(\psi)\Bigr\}\right]
=\Omega\!\left(\min\left\{1,\sqrt{\frac{T}{n}}\right\}\right).
\end{equation*}
In this construction the atomic excess coincides with the final-task excess, so the same bound holds for $\mathbb E_S[R_{\mathcal D}^{\theta}(F_{\widehat\psi})-\min_{\psi\in\Psi}R_{\mathcal D}^{\theta}(F_\psi)]$, and it is $\Omega(\sqrt{T/n})$ when $T\le n$.
\end{proposition}

\begin{proof}
Verify the construction of the statement. The weights of $P_\theta$ are nonnegative and sum to one; each $u^{(j)}$ has odd parity and each $v^{(j)}$ even parity, and each coordinate lies in exactly $2r+1$ of the $u^{(j)}$ and $2r$ of the $v^{(j)}$, so that for every coordinate $i$
\begin{equation}
P_\theta(X_i=1)=\frac12+\theta\gamma,
\qquad
P_\theta\!\left(\bigoplus_{i=1}^{T}X_i=1\right)=\frac12+\theta T\gamma .
\label{eq:parity_marginals}
\end{equation} The oracle performs serial XOR, so the final target is $p_T=\bigoplus_{i=1}^{T}X_i$. At atomic step $i$, $\psi_s$ incurs loss $\ell_{\mathrm{ok}}$ when $X_i=s$ and $\ell_{\mathrm{err}}$ otherwise; hence
\begin{equation*}
R_{\mathcal D_i}^{\theta}(\psi_s)
=\ell_{\mathrm{ok}}+\Delta_\ell P_\theta(X_i\ne s),
\qquad \Delta_\ell=\ell_{\mathrm{err}}-\ell_{\mathrm{ok}}=\log3 .
\end{equation*}
Writing $s_\theta=(1+\theta)/2$, Eq.~\eqref{eq:parity_marginals} gives
\begin{equation}
\sum_{i=1}^{T}
\left[
R_{\mathcal D_i}^{\theta}(\psi_{1-s_\theta})
-R_{\mathcal D_i}^{\theta}(\psi_{s_\theta})
\right]
=2\Delta_\ell T\gamma.
\label{eq:atomic_parity_excess}
\end{equation}
The learned rollout uses the same $s$ at every step. Since $T$ is odd, its final hard output is $s$, so the final loss is $\ell_{\mathrm{ok}}$ when $p_T=s$ and $\ell_{\mathrm{err}}$ otherwise, and
\begin{equation}
R_{\mathcal D}^{\theta}(F_{\psi_{1-s_\theta}})
-R_{\mathcal D}^{\theta}(F_{\psi_{s_\theta}})
=2\Delta_\ell T\gamma.
\label{eq:final_parity_excess}
\end{equation}
Equations~\eqref{eq:atomic_parity_excess} and~\eqref{eq:final_parity_excess} show that the atomic excess equals the final excess exactly.

For training, step $i$ uses only the $i$-th coordinate of its own draws, written $X_{i1},\dots,X_{in}$, so the $nT$ coordinates $X_{ij}$ used are independent. The sub-task ERM therefore chooses
\begin{equation*}
\widehat s
=\ind\!\left\{
\sum_{i=1}^{T}\sum_{j=1}^{n}X_{ij}\ge\frac{nT}{2}
\right\},
\end{equation*}
a majority test on $nT$ independent Bernoulli variables with mean $1/2+\theta\gamma$. The choice of $\gamma$ guarantees $0<\gamma\le1/(2T)$, and standard binomial anti-concentration gives a universal $c_1>0$ such that $P_\theta(\widehat s\ne s_\theta)\ge c_1$. Whenever the wrong hypothesis is selected, Eq.~\eqref{eq:atomic_parity_excess} gives atomic excess $2\Delta_\ell T\gamma$. Therefore
\begin{align*}
\mathbb E_{S}\!\left[
\sum_{i=1}^{T}R_{\mathcal D_i}^{\theta}(\psi_{\widehat s})
-\min_{\psi\in\Psi}\Bigl\{\sum_{i=1}^{T}R_{\mathcal D_i}^{\theta}(\psi)\Bigr\}
\right]
&\ge 2c_1\Delta_\ell T\gamma=\Omega\!\left(\min\left\{1,\sqrt{\frac{T}{n}}\right\}\right).
\end{align*}
By Eq.~\eqref{eq:final_parity_excess} the atomic excess equals the final-task excess here, so the same bound holds for $\mathbb E_{S}[R_{\mathcal D}^{\theta}(F_{\psi_{\widehat s}})-\min_{\psi\in\Psi}R_{\mathcal D}^{\theta}(F_\psi)]$; it is $\Omega(\sqrt{T/n})$ when $T\le n$.
\end{proof}
\paragraph{What the lower bound certifies.} The quantity bounded below is exactly the sample-dependent atomic excess of the upper bound---the sub-task ERM's excess over the best in-class value---and it coincides with the final-task excess. Since the per-step best-in-class loss is $\Theta(1)$ ($\ell_{\mathrm{ok}}+\tfrac12\Delta_\ell\approx0.837$), the excess is identifiable in this instance. Hence the $\Omega(\sqrt{T/n})$ rate is not an artifact of estimation: it concerns the estimation component only and says nothing about the approximation term.
\section{Full Experiments}
\label{app:full_exp}

In this section, we provide the extended experimental details and additional results that complement the main text.

\subsection{Atom-Task Experiments}
\label{subsec:aton_exp_app}

\subsubsection{Experimental Setup}
\label{subsubsec:aton_setup}

\paragraph{Atom-Task Benchmark.} To validate compositional reasoning capability, we introduce Atom-Task. This task requires the model to sequentially execute a series of atomic operations on a symbolic sequence multiple times, necessitating multi-step reasoning capabilities.

\textbf{Symbol System and Atomic Operations.} The input consists of sequences formed by 6 Greek letters $\{\alpha, \beta, \gamma, \delta, \varepsilon, \omega\}$. We define five basic operations (mapped to IDs t1--t5):
\begin{itemize}
    \item \textbf{Reverse (t1)}: Reverses the order of the sequence.
    \item \textbf{Substitute (t2)}: Swaps symbols in pairs ($\alpha \leftrightarrow \beta, \gamma \leftrightarrow \delta, \varepsilon \leftrightarrow \omega$).
    \item \textbf{Duplicate (t3)}: Duplicates each symbol consecutively.
    \item \textbf{Deduplicate (t4)}: Removes consecutive duplicates.
    \item \textbf{Rotate (t5)}: Moves the first element to the end.
\end{itemize}

\paragraph{Dataset Design and Distribution.}
A core challenge in compositional generalization is the distribution shift between training and testing. We design specific training datasets to investigate these effects.

\paragraph{Design Principles.}
Our training sets adhere to two core principles:
\begin{itemize}
    \item \textbf{Dominance of atomic primitives}: In Mixed 1-, 3-, and 5-Step, single-step samples constitute 80\% of the training trajectories, supporting the acquisition of basic operations.
    \item \textbf{Controlled exposure to composition}: Three- and five-step training chains provide structural patterns for compositional decomposition; in Mixed 3+5, the five-step chains use $\le2$ distinct operations, as specified in Section~\ref{sec:experiments}.
\end{itemize}

\textbf{Training Datasets.}
To evaluate OOD generalization, we design four training datasets:
\begin{enumerate}
    \item \textbf{Single-Step}: 5k uniformly distributed one-step samples.
    \item \textbf{Three-Step}: 1.66k three-step samples.
    \item \textbf{Mixed 3+5}: 1,333 samples (833 three-step and 500 five-step; the five-step chains use at most two distinct operations, bridging length and combinatorial generalization).
    \item \textbf{Mixed 1-, 3-, and 5-Step}: For each operation $op$, a dedicated 5k-sample dataset excludes $op$ from all multi-step chains but keeps it in single-step data. Specifically, it contains 4,000 one-step, 600 three-step, and 400 five-step samples (5{,}000 training trajectories, or 7{,}800 atomic transitions in total; see Table~\ref{tab:ablation_dist}).
\end{enumerate}

\textbf{Balanced Supervision:}
While the sample sizes vary across settings, the number of atomic transitions is held roughly constant: the single-step, three-step and mixed 3+5 datasets each contain about 5,000 transitions ($5{,}000$, $1{,}666\times3$, and $833\times3+500\times5$), while the operation-exclusion dataset contains $7{,}800$ and is compared across excluded operations under that same budget, which controls for confounding variables.

\textbf{Test Sets.} 
We evaluate on four test sets with increasing chain lengths: 3-step, 5-step, 7-step, and 9-step. Each set contains 500 unseen samples, requiring the model to perform compositional reasoning not explicitly memorized from training.

\begin{table}[ht!]
\centering
\caption{Sample distribution in Ablation Datasets. Each dataset excludes a specific operation from multi-step chains. The 5{,}000 trajectories comprise 7{,}800 atomic transitions.}
\label{tab:ablation_dist}
\footnotesize
\begin{adjustbox}{max width=\linewidth}
\begin{tabular}{@{}lcccc@{}}
\toprule
\textbf{Data} & \textbf{Total} & \textbf{Single-step} & \textbf{3-step} & \textbf{5-step} \\
\midrule
\texttt{Ablation\_no\_*} & 5,000 & 4,000 & 600 & 400 \\
\bottomrule
\end{tabular}
\end{adjustbox}
\end{table}

\subsubsection{Implementation Details}
\label{subsubsec:aton_impl}

For the Atom-Task, we use Qwen3-4B as the base model. Fine-tuning is performed using low-rank adaptation (LoRA; rank $16$, alpha $32$) for 3 epochs with a learning rate of $5\times10^{-5}$. The maximum generation length (\texttt{max\_new\_tokens}) is set to 2048 to accommodate longer reasoning chains. All experiments are conducted on a single NVIDIA RTX 4090 GPU.

\textbf{Parameter Efficiency:}
We employ LoRA with the following configuration:
\begin{itemize}
    \item Rank $r=16$ for a balance between expressiveness and parameter efficiency.
    \item LoRA modules are applied to all linear layers within the attention and MLP blocks.
\end{itemize}

We employ a Chain-of-Thought training format, where the model is trained to explicitly predict the intermediate output after each atomic operation. For the ``No-CoT'' baseline, the model is trained to predict only the final output without intermediate states. 

\textbf{Reproducibility:} For the main results (Table~\ref{tab:main_results_combined}), both CoT and No-CoT entries are means across 3 random seeds (42, 2026, 12345), and the same 3 seeds are used for the operation-exclusion study, the format ablation, the noise tests, and the scaling points. For the format ablation (Table~\ref{tab:format_ablation_merged}), the ablated CoT entries are means over 3 seeds and the original entries average the three-seed means of Table~\ref{tab:ablation_combined}. Standard deviations are reported as subscripts and the seed setting is stated in each caption; differences of a few points between entries sharing a seed set should be read as indicative rather than as significance statements.

\subsubsection{Details on Scaling Laws}
\label{subsubsec:scaling_details}

To complement Figure~\ref{fig:scaling}, we provide the detailed accuracy numbers for training-sample scaling and inference-example scaling in Tables~\ref{tab:n_scaling} and~\ref{tab:context_examples}, respectively. All scaling experiments are conducted using the ``Three-Step'' training dataset and evaluated on held-out test sets, each reported point being a mean over 3 seeds (42, 2026, 12345).

\begin{table}[ht!]
    \centering
    \renewcommand{\arraystretch}{1.1} %

    \begin{minipage}{0.48\textwidth}
        \centering
        \caption{Accuracy (\%) for Sample Size Scaling ($n$). Each entry is a mean over 3 seeds (42, 2026, 12345).}
        \label{tab:n_scaling}
\footnotesize
        \begin{adjustbox}{max width=\linewidth}
\begin{tabular}{lccc}
            \toprule
            \textbf{Prop.} & \textbf{3-step} & \textbf{5-step} & \textbf{7-step} \\
            \midrule
            \textbf{10\%}  & $91.93_{\mbox{\tiny$\pm 1.32$}}$ & $78.00_{\mbox{\tiny$\pm 1.18$}}$ & $58.73_{\mbox{\tiny$\pm 2.81$}}$ \\
            \textbf{20\%}  & $95.33_{\mbox{\tiny$\pm 0.41$}}$ & $84.27_{\mbox{\tiny$\pm 1.27$}}$ & $65.87_{\mbox{\tiny$\pm 2.33$}}$ \\
            \textbf{50\%}  & $98.13_{\mbox{\tiny$\pm 0.38$}}$ & $91.40_{\mbox{\tiny$\pm 1.30$}}$ & $80.40_{\mbox{\tiny$\pm 3.56$}}$ \\
            \textbf{100\%} & $99.33_{\mbox{\tiny$\pm 0.25$}}$ & $95.73_{\mbox{\tiny$\pm 0.68$}}$ & $88.87_{\mbox{\tiny$\pm 1.54$}}$ \\
            \bottomrule
        \end{tabular}
\end{adjustbox}
    \end{minipage}
    \hfill %
    \begin{minipage}{0.48\textwidth}
        \centering
        \caption{Accuracy (\%) versus Number of Inference-Time Examples. Each entry is a mean over 3 seeds (42, 2026, 12345).}
        \label{tab:context_examples}
\footnotesize
        \begin{adjustbox}{max width=\linewidth}
\begin{tabular}{lccc}
            \toprule
            \textbf{Shots} & \textbf{3-step} & \textbf{5-step} & \textbf{7-step} \\
            \midrule
            \textbf{0-shot} & $99.27_{\mbox{\tiny$\pm 0.09$}}$ & $93.40_{\mbox{\tiny$\pm 2.83$}}$ & $82.53_{\mbox{\tiny$\pm 1.76$}}$ \\
            \textbf{1-shot} & $99.33_{\mbox{\tiny$\pm 0.25$}}$ & $95.73_{\mbox{\tiny$\pm 0.68$}}$ & $88.87_{\mbox{\tiny$\pm 1.54$}}$ \\
            \textbf{2-shot} & $99.33_{\mbox{\tiny$\pm 0.09$}}$ & $96.60_{\mbox{\tiny$\pm 0.16$}}$ & $90.93_{\mbox{\tiny$\pm 1.32$}}$ \\
            \textbf{3-shot} & $99.07_{\mbox{\tiny$\pm 0.19$}}$ & $95.73_{\mbox{\tiny$\pm 0.66$}}$ & $89.93_{\mbox{\tiny$\pm 0.81$}}$ \\
            \bottomrule
        \end{tabular}
\end{adjustbox}
    \end{minipage}
\end{table}

\subsubsection{Extended Results and Analysis}
\label{subsubsec:aton_results_app}

\paragraph{Robustness to Noise in Reasoning Steps.}
To evaluate the robustness of CoT training to data quality, we conduct a noise interference experiment on the Qwen3-4B model. We test: \textbf{Multi-step (Pure 3-step)}, \textbf{Multi-step (Mixed 3+5)}, and the \textbf{w/o Substitute} ablation setting. (Note: These earlier noise robustness experiments used \texttt{max\_new\_tokens=1024}, but the conclusions remain consistent).

\textbf{Noise Injection Strategy.} We define the basic unit of noise injection as a single inference step. We select a proportion $p \in \{0.05, 0.1, 0.2\}$ of the total reasoning steps in the training set for interference. A standard reasoning step contains four discrete elements: the operation, the input state, the output state, and the remaining tasks. For each selected step, we randomly choose \textbf{two} of these four elements and perform either \textit{Tampering} (replacing content with random symbols) or \textit{Deletion} (emptying the content).

\begin{table}[ht!]
\centering
\caption{Each entry is a mean over 3 seeds (42, 2026, 12345). Robustness to training noise (Qwen3-4B). Accuracy (\%) under varying noise ratios. Each cell shows the result for the 3-step / 5-step / 7-step test sets.}
\label{tab:noise_robust}
\footnotesize
\begin{adjustbox}{max width=\linewidth}
\resizebox{\linewidth}{!}{%
\begin{tabular}{@{}lccc@{}}
\toprule
\textbf{Noise Ratio} & \textbf{Multi-step (Pure 3-step)} & \textbf{Multi-step (Mix$_{3+5}$)} & \textbf{w/o Substitute} \\
\midrule
$p=0.05$ & $99.73_{\mbox{\tiny$\pm 0.09$}}$ / $97.80_{\mbox{\tiny$\pm 0.86$}}$ / $90.47_{\mbox{\tiny$\pm 2.46$}}$ & $99.33_{\mbox{\tiny$\pm 0.09$}}$ / $98.53_{\mbox{\tiny$\pm 0.34$}}$ / $94.87_{\mbox{\tiny$\pm 0.81$}}$ & $96.20_{\mbox{\tiny$\pm 0.65$}}$ / $92.33_{\mbox{\tiny$\pm 1.06$}}$ / $85.47_{\mbox{\tiny$\pm 2.80$}}$ \\
$p=0.10$ & $99.47_{\mbox{\tiny$\pm 0.09$}}$ / $97.20_{\mbox{\tiny$\pm 0.59$}}$ / $91.67_{\mbox{\tiny$\pm 0.25$}}$ & $99.27_{\mbox{\tiny$\pm 0.34$}}$ / $98.87_{\mbox{\tiny$\pm 0.25$}}$ / $94.80_{\mbox{\tiny$\pm 0.49$}}$ & $95.80_{\mbox{\tiny$\pm 0.33$}}$ / $92.33_{\mbox{\tiny$\pm 1.27$}}$ / $86.73_{\mbox{\tiny$\pm 1.04$}}$ \\
$p=0.20$ & $99.40_{\mbox{\tiny$\pm 0.00$}}$ / $96.60_{\mbox{\tiny$\pm 0.75$}}$ / $90.73_{\mbox{\tiny$\pm 1.89$}}$ & $99.13_{\mbox{\tiny$\pm 0.09$}}$ / $98.93_{\mbox{\tiny$\pm 0.62$}}$ / $94.80_{\mbox{\tiny$\pm 0.28$}}$ & $96.93_{\mbox{\tiny$\pm 0.77$}}$ / $92.47_{\mbox{\tiny$\pm 0.50$}}$ / $87.13_{\mbox{\tiny$\pm 1.97$}}$ \\
\bottomrule
\end{tabular}
}
\end{adjustbox}
\end{table}

Table~\ref{tab:noise_robust} presents accuracy for 3-step / 5-step / 7-step test sets respectively. The results indicate strong robustness to noise in reasoning steps. Across all training configurations, increasing the noise ratio does not lead to significant performance degradation; accuracy remains largely consistent with the clean baselines reported in previous tables. In some cases, minor variations are observed but without a clear negative trend, suggesting that the learned compositional reasoning patterns are resilient to localized inconsistencies in the training data.

\subsection{GSM8K Experiments}
\label{subsec:gsm8k_exp}

To validate our findings on real-world, complex reasoning scenarios, we conduct experiments on the GSM8K benchmark using Llama2-7B and Qwen2.5-3B fine-tuned on MetaMathQA, a math instruction-tuning dataset (395K samples in total). This protocol evaluates a controlled within-family distribution shift from MetaMathQA to GSM8K rather than source-disjoint transfer. Due to computational constraints, we use 30,000 training samples for each experiment. For standard experiments (baseline and noise), we randomly sample these 30,000 instances from the full dataset; for distribution shift experiments, we select 30,000 samples according to specific shift criteria. %

\subsubsection{Experimental Setup}

\paragraph{Models and Training.} We employ Llama2-7B and Qwen2.5-3B as base models. Models are fine-tuned using LoRA (rank=16, alpha=16) for 1 epoch with a learning rate of \(2\times10^{-5}\) and a maximum sequence length of 512. Training data is drawn from the MetaMathQA dataset. For baseline and noise experiments, we randomly sample 30,000 instances; for distribution shift experiments, we construct 30,000-sample subsets by filtering according to three shift indicators (Topic, Complexity, Answer Scale). %
All experiments are conducted on a single NVIDIA RTX 4090 GPU, and each experiment is run with three random seeds (42, 2026, 12345); the training data is identical across seeds, so reported standard deviations reflect model-side randomness.

\paragraph{Evaluation.} All models are evaluated on the standard GSM8K test set. We report accuracy as the primary metric.

\subsubsection{CoT vs. No-CoT \& OOD Generalization}

\paragraph{Baseline Comparison.}
As shown in Table~\ref{tab:gsm8k_baseline}, CoT training significantly boosts accuracy over untrained baselines, while No-CoT training yields only marginal improvements or even degradation.

\begin{table}[t]
\centering
\begin{minipage}[t]{0.48\textwidth}
\centering
\caption{GSM8K baseline and comparative results. Each entry is a mean over 3 seeds (42, 2026, 12345).}
\label{tab:gsm8k_baseline}
\footnotesize
\begin{tabular}{@{}lll@{}}
\toprule
\textbf{Model} & \textbf{Experiment} & \textbf{Acc(\%)} \\
\midrule
\multirow{3}{*}{\textbf{Llama2-7B}} & Untrained & 2.50 \\
& CoT Training & $39.14_{\mbox{\tiny$\pm 0.99$}}$ \\
& No-CoT Training & $8.16_{\mbox{\tiny$\pm 2.50$}}$ \\
\midrule
\multirow{3}{*}{\textbf{Qwen2.5-3B}} & Untrained & 64.44 \\
& CoT Training & $79.23_{\mbox{\tiny$\pm 0.25$}}$ \\
& No-CoT Training & $19.41_{\mbox{\tiny$\pm 1.17$}}$ \\
\bottomrule
\end{tabular}
\end{minipage}\hfill
\begin{minipage}[t]{0.48\textwidth}
\centering
\caption{GSM8K accuracy under distribution shifts. Each entry is a mean over 3 seeds (42, 2026, 12345).}
\label{tab:gsm8k_shift}
\footnotesize
\begin{tabular}{@{}lll@{}}
\toprule
\textbf{Model} & \textbf{Shift Type} & \textbf{Acc(\%)} \\
\midrule
\multirow{3}{*}{\textbf{Llama2-7B}} & Scale Shift & $38.74_{\mbox{\tiny$\pm 0.70$}}$ \\
& Complexity Shift & $36.57_{\mbox{\tiny$\pm 0.17$}}$ \\
& Topic Shift & $37.51_{\mbox{\tiny$\pm 2.14$}}$ \\
\midrule
\multirow{3}{*}{\textbf{Qwen2.5-3B}} & Topic Shift & $78.60_{\mbox{\tiny$\pm 0.55$}}$ \\
& Complexity Shift & $79.53_{\mbox{\tiny$\pm 0.45$}}$ \\
& Answer Scale Shift & $79.10_{\mbox{\tiny$\pm 0.25$}}$ \\
\bottomrule
\end{tabular}
\end{minipage}
\end{table}

\paragraph{Distribution Shift Experiments.}
In contrast to the random sampling used for baseline and noise studies, here we deliberately create distribution gaps by filtering 30,000 samples from MetaMathQA based on the following criteria: %
\begin{itemize}
\item \textit{Topic Shift (ood\_domain)}: $\text{shift}_{\text{topic}} = \frac{|K_{\text{sample}} \setminus K_{\text{GSM8K}}|}{|K_{\text{sample}} \cup K_{\text{GSM8K}}|}$, where $K_{\text{sample}}$ is the keyword set of the sample and $K_{\text{GSM8K}}$ the top-20 keyword set of GSM8K; this asymmetric keyword-novelty score measures the fraction of sample keywords absent from GSM8K's set, normalized by the union size.
\item \textit{Complexity Shift (ood\_complexity)}: $\text{shift}_{\text{complexity}} = \frac{|\text{steps}_{\text{sample}} - \mu_{\text{GSM8K}}|}{\sigma_{\text{GSM8K}}}$ (absolute Z-score of the sample's step count relative to GSM8K's mean and standard deviation).
\item \textit{Answer Scale Shift (ood\_magnitude)}: $\text{shift}_{\text{magnitude}} = 1 - P_{\text{GSM8K}}(\text{magnitude}_{\text{sample}})$ (one minus the GSM8K probability of the sample's answer magnitude bin).
\end{itemize}
The results in Table~\ref{tab:gsm8k_shift} show that performance remains robust across all shifts, with only mild fluctuations. This stability suggests that the compositional patterns learned via CoT can generalize to practical domains like mathematical reasoning.

\paragraph{Robustness to Reasoning Noise.}
We further test resilience to imperfect supervision by training on 30,000 randomly sampled CoT instances from MetaMathQA, injected with noise via a controlled strategy: %
(1) \textit{Sample-level probability}: Each sample has a probability $p$ of being selected; (2) \textit{Single-step injection}: For selected samples, noise is injected into a single randomly chosen reasoning step; (3) \textit{Noise type}: `delete' (removing the step content) or `wrong\_number' (modifying the numerical result), applied with equal probability (50\%). This assesses the model's ability to handle corrupted intermediate reasoning.

\begin{table}[htbp]
\centering
\caption{Each entry is a mean over 3 seeds (42, 2026, 12345). GSM8K accuracy under training with different proportions ($p$) of reasoning noise. All models trained on 30,000 samples.}
\label{tab:gsm8k_noise}
\footnotesize
\begin{adjustbox}{max width=\linewidth}
\begin{tabular}{lccccc}
\toprule
\textbf{Model} & \textbf{$p=0$ (Baseline)} & \textbf{$p=0.05$} & \textbf{$p=0.10$} & \textbf{$p=0.20$} & \textbf{$p=0.50$} \\
\midrule
\textbf{Llama2-7B} & $39.14_{\mbox{\tiny$\pm 0.99$}}$ & $39.20_{\mbox{\tiny$\pm 0.56$}}$ & $39.73_{\mbox{\tiny$\pm 1.02$}}$ & $37.63_{\mbox{\tiny$\pm 0.88$}}$ & $37.66_{\mbox{\tiny$\pm 0.89$}}$ \\
\textbf{Qwen2.5-3B} & $79.23_{\mbox{\tiny$\pm 0.25$}}$ & $79.27_{\mbox{\tiny$\pm 0.25$}}$ & $78.57_{\mbox{\tiny$\pm 0.37$}}$ & $77.20_{\mbox{\tiny$\pm 0.60$}}$ & $75.69_{\mbox{\tiny$\pm 0.56$}}$ \\
\bottomrule
\end{tabular}
\end{adjustbox}
\end{table}

Both models exhibit strong resistance to noise (Table~\ref{tab:gsm8k_noise}). Even with up to 50\% of samples corrupted, accuracy fluctuates within a narrow band and never collapses. This suggests that the model is relatively insensitive to localized noise in intermediate steps and can still absorb the correct logical patterns.

\subsubsection{Summary of Findings}
The GSM8K experiments reveal three main patterns: (1) CoT training substantially improves accuracy over untrained and No-CoT baselines, highlighting the benefit of intermediate step supervision; (2) models are robust to significant reasoning noise, maintaining stable performance even when training data contains deleted or incorrect steps, indicating that supervision remains effective even when individual steps are corrupted; (3) models generalize effectively across distribution shifts in topic, complexity, and answer scale, with only mild accuracy fluctuations, indicating that the reasoning patterns acquired via CoT transfer robustly to real-world mathematical domains.

\end{document}